\pdfoutput=1

\documentclass[11pt]{article}

\usepackage[final]{acl}

\usepackage{times}
\usepackage{latexsym,enumitem}
\usepackage{adjustbox}
\usepackage{array,amsmath}
\usepackage[T1]{fontenc}

\usepackage[utf8]{inputenc}

\usepackage{fancyhdr}

\usepackage{microtype}
\usepackage{svg}
\usepackage{inconsolata}
\usepackage{enumitem}

\usepackage{graphicx}
\usepackage{listings}

\title{\textit{Just a Scratch} \includegraphics[scale=0.08]{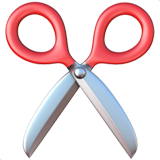}: Enhancing LLM Capabilities for Self-harm Detection through Intent Differentiation and Emoji Interpretation}

\author{
 \textbf{Soumitra Ghosh\textsuperscript{1\thanks{Authors contributed equally and are joint first authors.}}},
 \textbf{Gopendra Vikram Singh\textsuperscript{2\scriptsize*}},
 \textbf{Shambhavi\textsuperscript{2}},
 \textbf{Sabarna Choudhury\textsuperscript{3}},
\\
 \textbf{Asif Ekbal\textsuperscript{3}}
\\
\\
 \textsuperscript{1}Fondazione Bruno Kessler, Trento, Italy,
  \textsuperscript{2}Indian Institute of Technology Patna, India,\\
 \textsuperscript{3}Indian Institute of Technology Jodhpur, India\\
\{ghosh.soumitra2,gopendra.99\}@gmail.com, shambhavi\_2311ai38@iitp.ac.in,\\ \{m23aid067,asif\}@iitj.ac.in
}

\begin{document}
\thispagestyle{fancy} 

\maketitle

\begin{abstract}
Self-harm detection on social media is critical for early intervention and mental health support, yet remains challenging due to the subtle, context-dependent nature of such expressions. Identifying self-harm intent aids suicide prevention by enabling timely responses, but current large language models (LLMs) struggle to interpret implicit cues in casual language and emojis. This work enhances LLMs’ comprehension of self-harm by distinguishing intent through nuanced language–emoji interplay. We present the \textit{C}entennial \textit{E}moji \textit{S}ensitivity \textit{M}atrix (\textit{CESM-100})—a curated set of 100 emojis with contextual self-harm interpretations—and the \textit{S}elf-\textit{H}arm \textit{I}dentification a\textit{N}d intent \textit{E}xtraction with \textit{S}upportive emoji sensitivity (\textit{SHINES}) dataset, offering detailed annotations for self-harm labels, casual mentions (CMs), and serious intents (SIs). Our unified framework:
a) enriches inputs using CESM-100;
b) fine-tunes LLMs for multi-task learning—self-harm detection (primary) and CM/SI span detection (auxiliary);
c) generates explainable rationales for self-harm predictions. We evaluate the framework on three state-of-the-art LLMs—Llama 3, Mental-Alpaca, and MentalLlama—across zero-shot, few-shot, and fine-tuned scenarios. By coupling intent differentiation with contextual cues, our approach commendably enhances LLM performance in both detection and explanation tasks, effectively addressing the inherent ambiguity in self-harm signals. The \textit{SHINES} dataset, \textit{CESM-100} and codebase are publicly available at: \url{https://www.iitp.ac.in/%7eai-nlp-ml/resources.html#SHINES}
\end{abstract}

\section{Introduction}

\begin{table}[!ht]
\centering
\caption{Examples of social media posts illustrating the challenges in detecting self-harm. Spans highlighted in blue indicate \textit{Casual Mentions} and those in red indicate \textit{Serious Intent}.}
\label{tab:exmp}
\begin{adjustbox}{max width=0.47\textwidth}
\begin{tabular}{p{10cm}}
\hline
\textbf{Example Post and Challenge} \\
\hline
\textbf{Post: }I’ve had enough. I don’t even want to see tomorrow. Maybe I should just \textcolor{red}{end it all}.\includegraphics[scale=0.07]{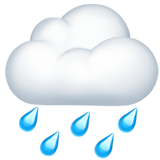} \includegraphics[scale=0.009]{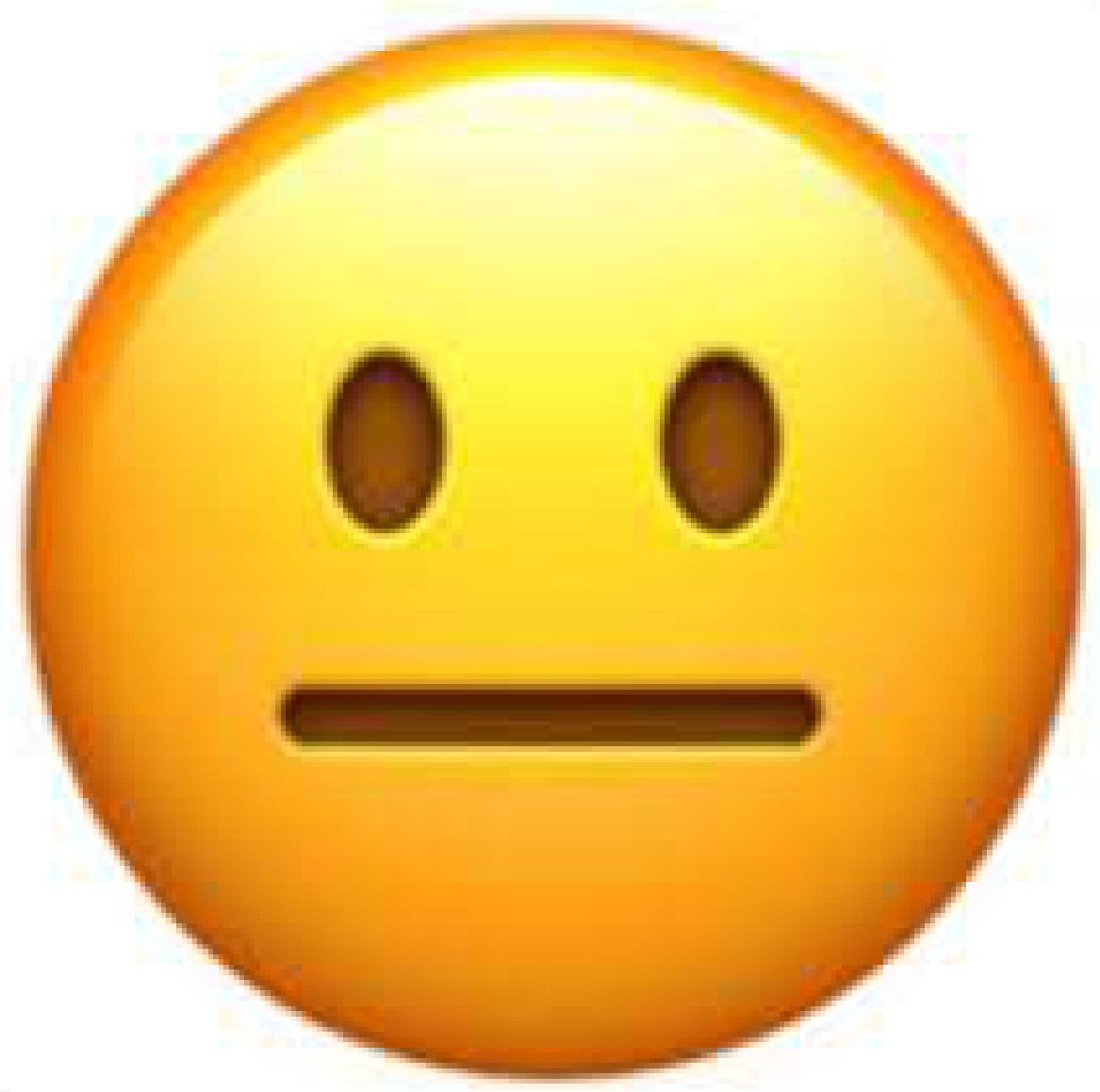} \\ \textbf{Challenge: }This post clearly indicates serious self-harm intent. The language used is direct and expresses a desire to end one's life. The emojis, a cloud with rain and a pensive face, reflect sadness and emotional distress, aligning with the serious tone of the text. \\ \hline 
\textbf{Post: }So tired of pretending everything is okay. Maybe it’s time to make some \textcolor{blue}{drastic changes... \includegraphics[scale=0.07]{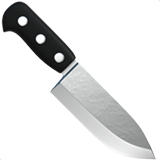} \includegraphics[scale=0.07]{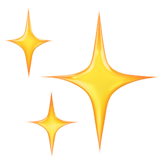}}  \\
\textbf{Challenge: }The knife emoji could suggest self-harm, while the sparkles emoji generally indicates positivity or hope. This mixed signal makes it ambiguous and might lead to misinterpretation by LLMs. \\ \hline 
\textbf{Post: }Just had a rough day. Feeling like I’m on the \textcolor{blue}{edge of losing it}. \includegraphics[scale=0.07]{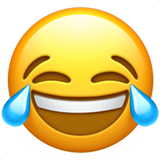} \includegraphics[scale=0.07]{images/knife.png} \\
\textbf{Challenge:} While the \includegraphics[scale=0.07]{images/knife.png} emoji could imply self-harm, the \includegraphics[scale=0.07]{images/joy.png} emoji suggests a light-hearted or ironic tone. ChatGPT-4o classifies as non self-harm whereas Google's GEMINI classifies as possible self-harm. \\
\hline
\end{tabular}
\end{adjustbox}

\end{table}

Self-harm is a pressing mental health concern, often serving as a coping mechanism for distress and, in some cases, a precursor to suicidal behaviors. Early identification of self-harm signals is crucial for timely intervention, making automated detection a valuable tool for suicide prevention and mental health monitoring. Social media platforms, where individuals frequently share their emotions, provide a unique opportunity to detect self-harm risk factors in real time. However, this task is inherently complex due to the informal, ambiguous, and multimodal nature of online discourse.

\emph{\textbf{Role of Language Nuance:}} While explicit self-harm statements are easier to detect, posts expressing distress through casual remarks, sarcasm, or irony, complicates automated detection. This complexity is illustrated in Table~\ref{tab:exmp}, which demonstrates the difficulty in distinguishing casual mentions from serious intent. Casual mention (CM) often use violent or distressing symbols hyperbolically, as seen in the last example. In contrast, serious intent (SI) involves genuine signs of distress, as shown in the first example. Recognizing the nuanced expressions of distress—such as sarcasm, irony, or casual remarks—is vital for effective self-harm detection. However, even advanced Large Language Models (LLMs) encounter significant challenges in accurately interpreting these subtleties. 
In our study, intent refers to the explicitness or implicitness of self-harm expressions, differing from intent detection in domains like search queries.



\emph{Emojis and Intent Differentiation:} Emojis play a crucial yet under-explored role in self-harm detection, as they can amplify, alter, or obscure textual meaning. For example, "I can't take this anymore \includegraphics[scale=0.06]{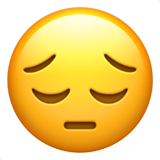}" signals emotional distress, whereas "I'm done with this \includegraphics[scale=0.06]{images/joy.png}" conveys a casual tone despite similar phrasing. Existing models often overlook such contextual variations, leading to misinterpretations. Addressing this gap requires a more nuanced approach that integrates both textual and emoji-based cues for improved comprehension.

The key contributions are summarized below:

\begin{itemize}[nolistsep]
\item Enhancing detection accuracy by explicitly addressing the nuanced differences between casual mentions (CM) and serious intent (SI) in self-harm expressions.
\item \textit{CESM-100}: Introducing a curated set of 100 emojis with contextual self-harm interpretations and intuitive attributes to augment LLMs’ multimodal understanding.
\item \textit{SHINES} Dataset: Proposing a novel dataset (\textbf{S}elf-\textbf{H}arm \textbf{I}dentification a\textbf{N}d intent \textbf{E}xtraction with \textbf{S}upportive emoji sensitivity) that includes self-harm annotations, CM and SI span labels, and emoji interpretations derived from \textit{CESM-100} for enhanced LLM training.
\item Developing an LLM fine-tuning framework for self-harm classification, CM/SI span extraction, and rationale generation to improve performance across tasks.
\item \textit{Comprehensive LLM Evaluation:} Assessing three state-of-the-art LLMs (Llama 3, Mental-Alpaca, and MentalLlama) under zero-shot, few-shot, single-task, and multi-task experimental settings to benchmark their effectiveness in self-harm detection.
\end{itemize}

\section{Related Work}
 Self-harm detection on social media is underexplored compared to stress detection, depression identification, and suicidal ideation. While self-harm correlates with suicidal ideation, literature indicates they are distinct phenomena with different behavioral, psychological, and clinical characteristics. For example, \citet{klonsky2013relationship} highlight that non-suicidal self-injury (NSSI) and suicidal behaviors differ in intent, frequency, and associated risk factors. Similarly, \citet{whitlock2013nonsuicidal} suggest that while self-harm can be a gateway to suicidal behaviors, it is not inherently indicative of suicidal ideation.

\textbf{Self-Harm in the Clinical Context: }
Self-harm, linked with depression, schizophrenia, and insomnia, is a significant concern in mental health \cite{lim2022investigating}. Clinical studies emphasize early detection to mitigate risks \cite{ennis1989depression}, while genetic research highlights the interplay of inherited traits and environmental factors \cite{campos2020genetic,russell2021exploration}. Despite these advancements, applying clinical insights to automated self-harm detection in digital spaces remains limited.

\textbf{Emojis, Mental Health, and Self-Harm: }
Emojis are emerging as key indicators of mental health, reflecting emotions and stress levels \cite{halverson2023content,grover2022exploiting,chan2022enhancing}. Their interpretations vary across individual and cultural contexts \cite{danesi2022emotional}, and tools like the Current Mood and Experience Scale incorporate emojis to assess well-being \cite{davies2024emoji}. However, their potential to distinguish casual from serious self-harm intent remains underexplored.

\textbf{LLMs and Mental Health: }
LLMs have shown promise in mental health applications, from automated diagnosis to therapeutic interventions. They have been employed in motivational interviewing \cite{welivita2023boosting}, cognitive behavioral therapy \cite{ding2022improving}, and simulating mental health support \cite{yu2024experimental,maddela2023training}. Despite their versatility, self-harm detection remains challenging due to the subtlety of expressions and multimodal nuances like emojis \cite{deshpande2021self}.  

A study by \citet{mcbain2025competency} evaluated the competency of LLMs like ChatGPT-4o, Claude 3.5 Sonnet, and Gemini 1.5 Pro in assessing appropriate responses to suicidal ideation. The findings revealed that these models often rated responses as more appropriate than expert suicidologists did, indicating an upward bias and a potential misinterpretation of the severity of distress signals. Similarly, research by \citet{grabb2024risks} highlighted that existing LLMs are insufficient in matching the standards provided by human professionals in mental health contexts. The study found that these models could cause harm if accessed during mental health emergencies, failing to protect users and potentially exacerbating existing symptoms. Furthermore, a study on the safety of LLMs \citep{heston2023safety} in addressing depression demonstrated that these models may not consistently detect and address hazardous psychological states. The research indicated that LLMs often postponed referrals to human support until severe depression levels were reached, potentially endangering users by not providing timely intervention.

Our work addresses this gap by leveraging LLMs to detect nuanced self-harm indicators, integrating contextual emoji analysis and tackling the distinction between casual mentions and serious intent in informal, multimodal online spaces.

\begin{table}[b!]
    \centering
        \caption{Dataset statistics.}
    \label{tab:post_data}
\begin{adjustbox}{max width=0.48\textwidth}    
    \begin{tabular}{l|r}
        \hline
        \textbf{Category} & \textbf{Value} \\
        \hline
        Total Posts & 5206 \\
        \hline
        Self-Harm Posts & 2499 \\
        \hline
        Non-Self-Harm Posts & 2707 \\
        \hline
        Posts with Emoji & 3067 \\
        \hline
        Posts without Emoji & 2139 \\
        \hline
        Average Length of Posts & 206 words \\
        \hline
        Self-Harm Posts with CM Spans & 34 \\
        \hline
        Self-Harm Posts with SI Spans & 2488 \\
        \hline
        Non Self-harm Posts with CM Spans & 2707 \\
        \hline
        Non Self-harm Posts with SI Spans & 0 \\
        \hline
    \end{tabular}
\end{adjustbox}    
\end{table}

\section{Dataset}

In this section, we outline the development of the \textbf{SHINES} dataset: \textbf{S}elf-\textbf{H}arm \textbf{I}dentification a\textbf{N}d intent \textbf{E}xtraction with \textbf{S}upportive emoji sensitivity. This dataset includes 5206 manually annotated social media posts with self-harm labels, CMs and SIs. It also incorporates emoji interpretations from \textbf{CESM-100}, a newly curated resource that we developed to provide contextualized emoji meanings, enhancing the dataset’s depth and utility.

\subsection{Data Collection}
Collecting self-harm posts, despite their prevalence on social media, is challenging due to their scarcity and dispersed nature. The rarity of explicit self-harm disclosures stems from social stigma and psychological barriers, while identifying posts with emojis adds complexity due to the variability and subtlety in emoji usage. We initially collected over 5k posts from mental health and self-harm subreddits, including \texttt{SuicideWatch}, \texttt{emotionalabuse}, \texttt{helpmecope}, \texttt{selfharm}, etc.\footnote{Full list is in Appendix ~\ref{a.1}}. After filtering\footnote{Filtering steps are in Appendix ~\ref{a:df}} out noisy texts—such as very short posts or those with titles but no body content—we were left with 4206 posts. We used Presidio\footnote{\url{https://github.com/microsoft/presidio}}, an open-source tool for detecting and anonymizing sensitive information, to remove personally identifiable information (PII).

\subsection{Data Annotation}
Three independent annotators\footnote{The annotation team includes: a postdoc at a foreign research lab with salary as per company norms, an assistant professor at a reputed university with a standard academic salary, and a contract-based computational linguist affiliated with a reputed institute’s CSE department.} assessed each post to determine whether it pertained to self-harm or non-self-harm, using majority voting for final labels. Fleiss' Kappa \cite{spitzer1967quantification} was calculated, yielding a score of 0.78 (0.82 for self-harm, 0.74 for non-self-harm), indicating substantial agreement.

Furthermore, each post was annotated for spans indicating casual mention and serious intent. Following methodologies by \citet{poria2021recognizing} and \citet{ghosh2022multitask}, up to three spans per category were identified. Span aggregation followed \citet{gui2018event}. Inter-rater agreement was evaluated using the macro-F1 metric, resulting in F1-scores of 0.66 for CM and 0.69 for SI, demonstrating robust annotation quality. Table~\ref{tab:post_data} summarizes key characteristics of the dataset.

\begin{table*}[!ht]
\centering
\caption{Snapshot of the \textit{Centennial Emoji Sensitivity Matrix (CESM-100)}. CM: Casual Mention, SI: Serious Intent.}
\begin{adjustbox}{max width=1\textwidth}
\setlength\tabcolsep{5pt}
\renewcommand{\arraystretch}{0.9}
\begin{tabular}{c|c|p{11.5cm}|c|c}
\hline
\textbf{Emoji} & \textbf{Usual Meaning} & \textbf{Contextual Meaning} & \textbf{CM Chance} & \textbf{SI Chance} \\
\hline
\includegraphics[scale=0.07]{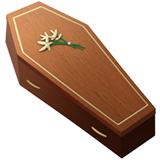} & Coffin & Indicates thoughts of death or suicide. & Low & High \\
\hline
\includegraphics[scale=0.07]{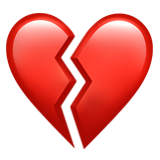} & Broken Heart & Represents intense emotional pain, often related to feelings of hopelessness. & Medium & High \\
\hline
\includegraphics[scale=0.07]{images/knife.png} & Kitchen Knife & Used to signify thoughts or acts of self-harm. & Low & High \\
\hline
\includegraphics[scale=0.012]{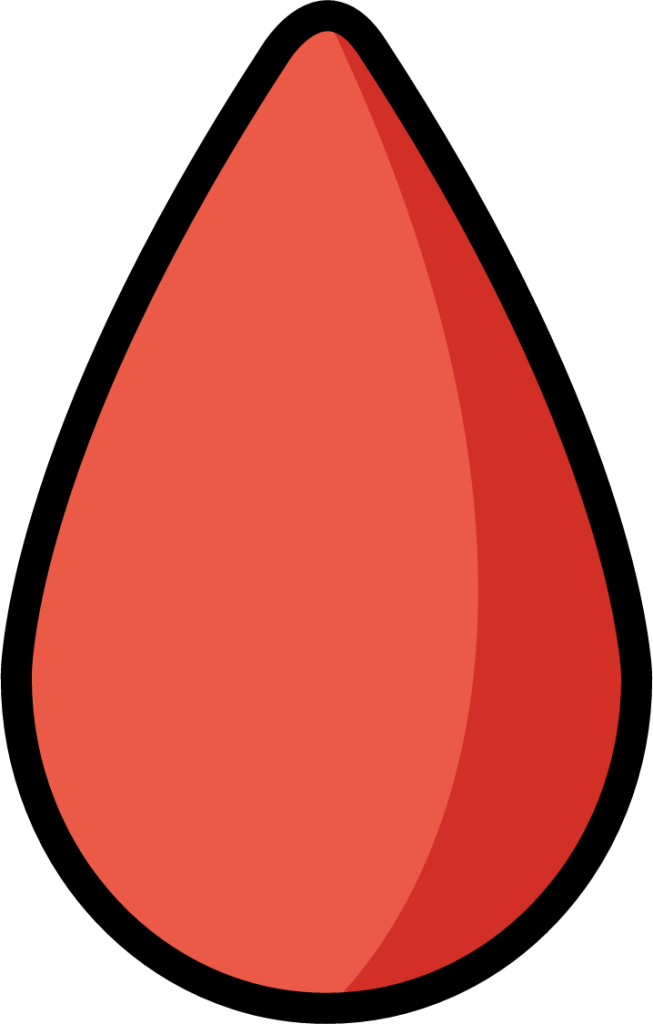} & Drop of Blood & Represents bleeding or injury, often associated with self-harm. & Medium & High \\
\hline
\includegraphics[scale=0.07]{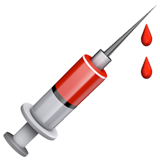} & Syringe & Often used in contexts related to injections or medication, can also relate to self-harm or medical issues. & Low & Medium \\
\hline
\includegraphics[scale=0.07]{images/joy.png} & Face with Tears of Joy & Often used to mask deeper emotional pain behind a facade of laughter. & High & Medium \\
\hline
\includegraphics[scale=0.07]{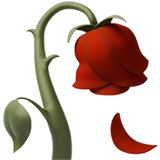} & Wilted Flower & Symbolizes decay or decline, often used metaphorically for sadness. & Medium & High \\
\hline
\includegraphics[scale=0.07]{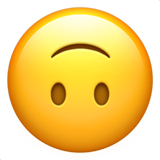} & Upside-Down Face & Represents hiding true emotions behind a facade of indifference or humor. & High & Medium \\
\hline
\end{tabular}
\end{adjustbox}
\label{tab:emoji_matrix}
\end{table*}

\subsection{Enhancing Dataset Robustness with Synthetic Posts}
To augment training data, we generated 1000 synthetic posts using ChatGPT-3.5\footnote{\url{https://chatgpt.com}} with few-shot prompting based on examples from our annotated dataset (see Table~\ref{tab:prompts for synthetic data}). This included 500 posts each for self-harm and non-self-harm categories, ensuring balance and deeper analysis. While ChatGPT generated fluent, realistic posts, its emoji usage often deviated from authentic patterns. Two annotators (A1, A2) manually revised emoji usage, dividing the posts equally.

To assess alignment with original data, annotator A3 mixed 500 original and 500 synthetic posts revised by A1, asking A2 to classify each as "original" or "synthetic". A1 performed the same task with posts revised by A2. The average F1 score of 58\%—above random but not definitive—demonstrates that synthetic posts closely resemble original ones, confirming their quality.

The synthetic posts thus generated were then added to the initial collection of 4206 posts. Our final dataset now consists of 5206 posts, including 4206 manually annotated original samples and 1000 validated synthetic samples, all annotated for self-harm/non self-harm classification as well as CM and SI spans. The SHINES CM/SI schema and CESM-100 emoji interpretations were validated by a psychiatrist with over 12 years of clinical experience, ensuring clinical rigor in capturing nuanced self-harm expressions. A subset of SHINES annotations is available for review, with the full dataset to be released pending research approval.

\section{Centennial Emoji Sensitivity Matrix (CESM-100)}
This emoji matrix aims to contextualize self-harm information within social media posts.

\subsection{Collection of Emoji Meanings}
A subset of emojis was extracted from the collected posts (as detailed in last section) related to self-harm and analyzed to understand their usage in these contexts. A broader set of emojis was also reviewed and included based on the collective expertise of the authors and annotators. This process led to the creation of the \textit{CESM-100}, a comprehensive collection of 100 emojis annotated with contextual information relevant to self-harm.

The usual meanings of emojis were sourced from \textit{Emojipedia}\footnote{\url{https://emojipedia.org/}}, a comprehensive resource for emoji definitions. Emojipedia provides detailed descriptions of each emoji's general usage across different platforms, which formed the basis for understanding their common meanings. The detailed guidelines for developing CESM-100 is presented in section~\ref{guidelines} of the appendix.

\subsection{Annotation Process}
\textbf{\emph{Casual Mention vs. Serious Intent:}}
Emojis were categorized based on their likelihood of appearing in Casual Mention (non-serious contexts) or Serious Intent (signifying emotional distress or self-harm). Three levels of association were defined:

\begin{itemize}[nolistsep]
    \item \textbf{Low}: Rarely used in the given context.
    \item \textbf{Medium}: Moderately associated.
    \item \textbf{High}: Frequently associated with the context.
\end{itemize}

Casual Mention reflects emojis used in non-serious, everyday contexts, while Serious Intent pertains to emojis used to signify deeper emotional distress or self-harm.

\textbf{\emph{Calculation of Inter-Annotator Agreement:}}
Inter-annotator agreement was assessed using Fleiss' Kappa, with scores of 0.71 for CM and 0.75 for SI, reflecting substantial agreement and consistency among annotators.

\subsection{Determining Contextual Meaning in Self-Harm}
Annotators collaboratively determined each emoji’s contextual meaning in self-harm discussions, focusing on its frequency and role in conveying distress or emotional pain. The final matrix (Table~\ref{tab:emoji_matrix}) details each emoji, its typical meaning, its contextual interpretation in self-harm, and its association with Casual Mention or Serious Intent. Table~\ref{tab:post_example} presents an annotated SHINES dataset sample with spans for casual mention (CM), serious intent (SI), and Self-harm/Non Self-harm labels. Guidelines for generating emoji interpretations relevant to self-harm are in section~\ref{guidelines} of the Appendix.

\begin{table}[h!]
    \centering
    \caption{Example post from the \textbf{SHINES} dataset.}
\begin{adjustbox}{max width=0.45\textwidth}    
\renewcommand{\arraystretch}{1}
    \begin{tabular}{p{1.6cm}|p{6.7cm}}
        \hline
        \textbf{Field} & \textbf{Value} \\
        \hline
        \textbf{\small Post Text} & Title: Venting about Life's Frustrations with a Dash of Dark Humor \includegraphics[scale=0.05]{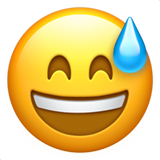} \\
        & Post: Ugh, just spilled coffee all over my keyboard... might as well electrocute myself next! Just one of those days, am I right? \\
        & Remember, it's all in good fun! Stay safe, friends. \includegraphics[scale=0.05]{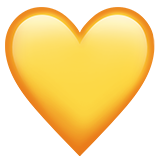} \\
        \hline
        \textbf{Label} & Non Self-harm \\
        \hline
        \textbf{CM Spans} & electrocute myself next \\
        \hline
        \textbf{SI Spans} & None \\
        \hline
        \textbf{Emojis} & 
        \begin{tabular}{ll}
            \textbullet & \textbf{Emoji}: \includegraphics[scale=0.05]{images/sweat_smile.png} \\
            & \small \textbf{Usual Meaning}: Smiling Face with Sweat \\
            & \textbf{Contextual Meaning}: Indicates \\
            & nervousness or awkwardness, often \\
            & masking deeper emotional pain. \\
            & \textbf{Casual Mention Chance}: High \\
            & \textbf{Serious Intent Chance}: Medium \\
            \textbullet & \textbf{Emoji}: \includegraphics[scale=0.05]{images/yellow_heart.png} \\
            & \textbf{Usual Meaning}: Yellow Heart \\
            & \textbf{Contextual Meaning}: Indicates \\
            & friendship or warmth, which could be \\
            & used to counteract feelings of distress. \\
            & \textbf{Casual Mention Chance}: High \\
            & \textbf{Serious Intent Chance}: Low \\
        \end{tabular} \\
        \hline
    \end{tabular}
    \end{adjustbox}
    \label{tab:post_example}
\end{table}

\section{Emoji Representation in Self-Harm and Non-Self-Harm Discourse}
Inspired by recent studies \cite{cohn2019grammar,yang2024elco}, we investigated emoji occurrences as single characters (\includegraphics[scale=0.13]{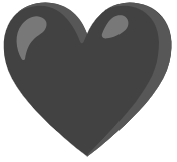}) or in compositions (\includegraphics[scale=0.07]{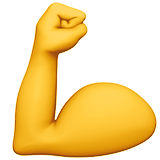} \includegraphics[scale=0.07]{images/sparkles.png}) across self-harm and non-self-harm posts (ref. Table~\ref{tab:emojicomp}). The table reveals two key trends in emoji usage across self-harm (SH) and non-self-harm (NSH) posts. Single-emoji compositions remain the most frequent overall, with a significantly higher presence in SH posts compared to NSH posts, supporting prior findings that visual symbols aid emotional expression in sensitive discussions \cite{cohn2019grammar}. However, as emoji composition becomes more complex, SH posts increasingly dominate, suggesting a preference for multi-emoji expressions in self-harm contexts.

\begin{table}[!ht]
\centering
\caption{Emoji Composition Distribution over Self Harm (SH) and Non-Self-Harm (NSH) posts.}
\label{tab:emojicomp}
\begin{adjustbox}{max width=0.4\textwidth}    
\begin{tabular}{c|c|c|c}
\hline
Count & Composition Example & SH & NSH \\
\hline
1 & \includegraphics[scale=0.13]{images/black_heart.png} & 7815 & 5359 \\
\hline
2 & \includegraphics[scale=0.07]{images/knife.png} \includegraphics[scale=0.012]{images/drop-of-blood.653x1024.png} & 2032 & 116 \\
\hline
3 & \includegraphics[scale=0.13]{images/black_heart.png} \includegraphics[scale=0.07]{images/muscle.png} \includegraphics[scale=0.07]{images/sparkles.png}  & 802 & 36 \\
\hline
4+ & \includegraphics[scale=0.07]{images/pensive.png} \includegraphics[scale=0.19]{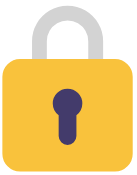}   \includegraphics[scale=0.07]{images/knife.png} \includegraphics[scale=0.07]{images/broken_heart.png}  & 82 & 22 \\
\hline

\end{tabular}
\end{adjustbox}
\end{table}

\begin{table}[!ht]
\centering
\caption{Frequency Distribution of Emojis by Category and Intent Type. DR: Direct Representation, MU: Metaphorical Use, SL: Semantic List.}
\label{tab:emoji_freq}
\begin{adjustbox}{max width=0.35\textwidth}    
\begin{tabular}{l|c|c|c}
\hline
\textbf{Intent Type} & \textbf{DR} & \textbf{MU} & \textbf{SL} \\
\hline
Serious Intent & 2268 & 2761 & 1913 \\
Casual Mention & 1668 & 1479 & 608 \\
\hline
\end{tabular}
\end{adjustbox}
\end{table}

Additionally, we present the distribution of emojis in the context of casual mention and serious intent over three predominant compositional strategies \cite{yang2024elco} within our \textit{SHINES} dataset: Direct Representation, Metaphorical Representation, and Semantic List (ref. Table~\ref{tab:emoji_freq}). Serious Intent posts favor metaphorical emoji usage over direct representation, suggesting a preference for symbolic expression in conveying distress. In contrast, Casual Mentions exhibit the opposite trend, where emojis are used more literally. Across both categories, the semantic list has the lowest usage, but it remains notably present in Serious Intent posts, indicating some structured expression of distress. These patterns highlight distinct expressive functions of emojis, with metaphorical use being central to serious emotional states, while direct representation dominates casual references. Table~\ref{tab:emoji_freq_cmsi} lists representative emojis for casual mentions and serious intent, sorted by overall frequency in the dataset. 

\begin{table}[!ht]
\centering
\caption{5 Most Frequently Occurring Emojis (non-exhaustive but representative) within the context of Casual Mentions (CMs) and Serious Intent (SI).}
\label{tab:emoji_freq_cmsi}
\begin{adjustbox}{max width=0.4\textwidth}    
\begin{tabular}{l|c|c}
\hline
\textbf{Emoji} & \textbf{CM Frequency} & \textbf{SI Frequency} \\
\hline
\multicolumn{3}{c}{\small Sorted with respect to occurrences in context of SI} \\
\hline
\includegraphics[scale=0.05]{images/broken_heart.png} & 73 & 1227 \\
\includegraphics[scale=0.05]{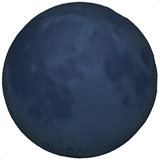} & 34 & 1083 \\
\includegraphics[scale=0.05]{images/muscle.png} & 43 & 1071 \\
\includegraphics[scale=0.05]{images/pensive.png} & 41 & 1023 \\
\includegraphics[scale=0.07]{images/sparkles.png} & 30 & 830 \\
\hline
\multicolumn{3}{c}{\small Sorted with respect to occurrences in context of CMs} \\
\hline
\includegraphics[scale=0.05]{images/sweat_smile.png} & 706 & 2 \\
\includegraphics[scale=0.05]{images/joy.png} & 424 & 0 \\
\includegraphics[scale=0.004]{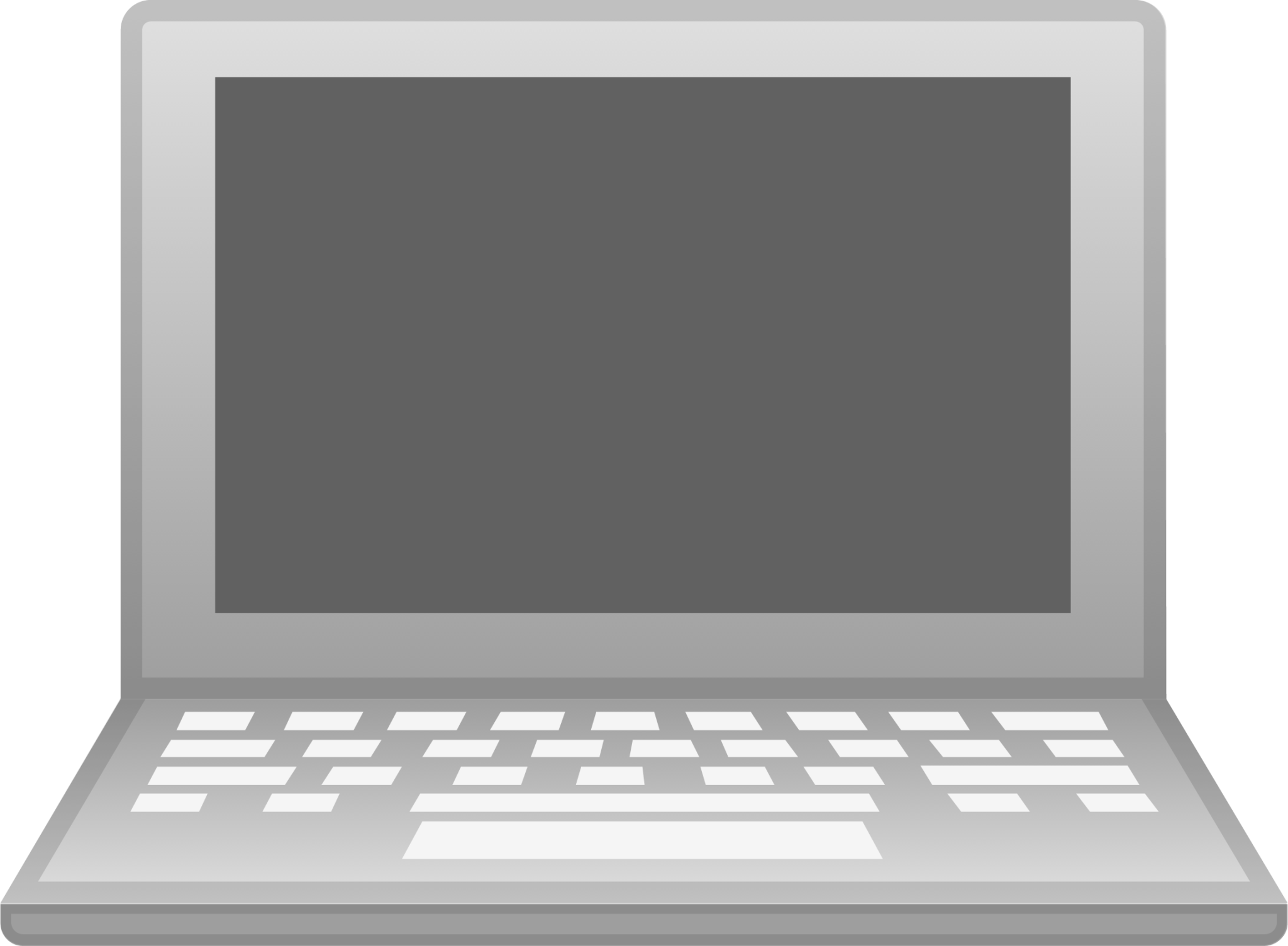} & 349 & 0 \\
\includegraphics[scale=0.07]{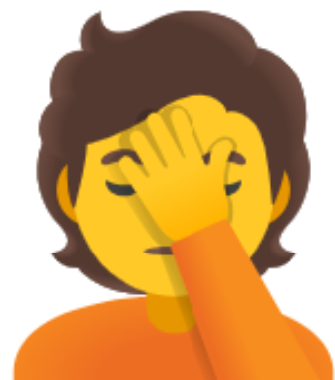} & 218 & 0 \\
\includegraphics[scale=0.02]{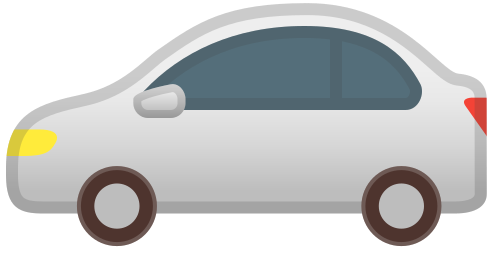} & 181 & 1 \\
\hline
\end{tabular}
\end{adjustbox}
\end{table}

Table~\ref{tab:emoji_freq_shnsh} depicts the most frequently occurring emojis within self-harm and non-self-harm posts in the SHINES dataset. Insights from the qualitative analysis of emoji usage in self-harm expressions is presented in Section~\ref{App:Emojianalysis} of the appendix.

\begin{table}[!ht]
\centering
\caption{5 Most Frequently Occurring Emojis (non-exhaustive but representative) within the context of Self Harm (SH) and Non-Self-Harm (NSH).}
\label{tab:emoji_freq_shnsh}
\begin{adjustbox}{max width=0.4\textwidth}    
\begin{tabular}{l|c|c}
\hline
\textbf{Emoji} & \textbf{SH Frequency} & \textbf{NSH Frequency} \\
\hline
\multicolumn{3}{c}{\small Sorted with respect to occurrences in context of SH} \\
\hline
\includegraphics[scale=0.05]{images/broken_heart.png} & 1337 & 56 \\
\includegraphics[scale=0.05]{images/new_moon.png} & 1122 & 11\\
\includegraphics[scale=0.05]{images/muscle.png} & 1110 & 19 \\
\includegraphics[scale=0.05]{images/pensive.png} & 1041 & 19\\
\includegraphics[scale=0.07]{images/sparkles.png} & 842 & 9\\
\hline
\multicolumn{3}{c}{\small Sorted with respect to occurrences in context of NSH} \\
\hline
\includegraphics[scale=0.05]{images/sweat_smile.png} & 3 & 721 \\
\includegraphics[scale=0.05]{images/joy.png} & 0 & 445 \\
\includegraphics[scale=0.004]{images/laptop-computer-emoji-2048x1504-up6ytfor.png} & 0 & 368 \\
\includegraphics[scale=0.05]{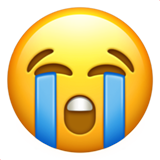} & 17 & 291 \\
\includegraphics[scale=0.08]{images/face_palm.png} & 0 & 233 \\
\hline
\end{tabular}
\end{adjustbox}
\end{table}

\section{Methodology}
\begin{figure}[!ht]
\centerline{\includegraphics[scale=0.39]{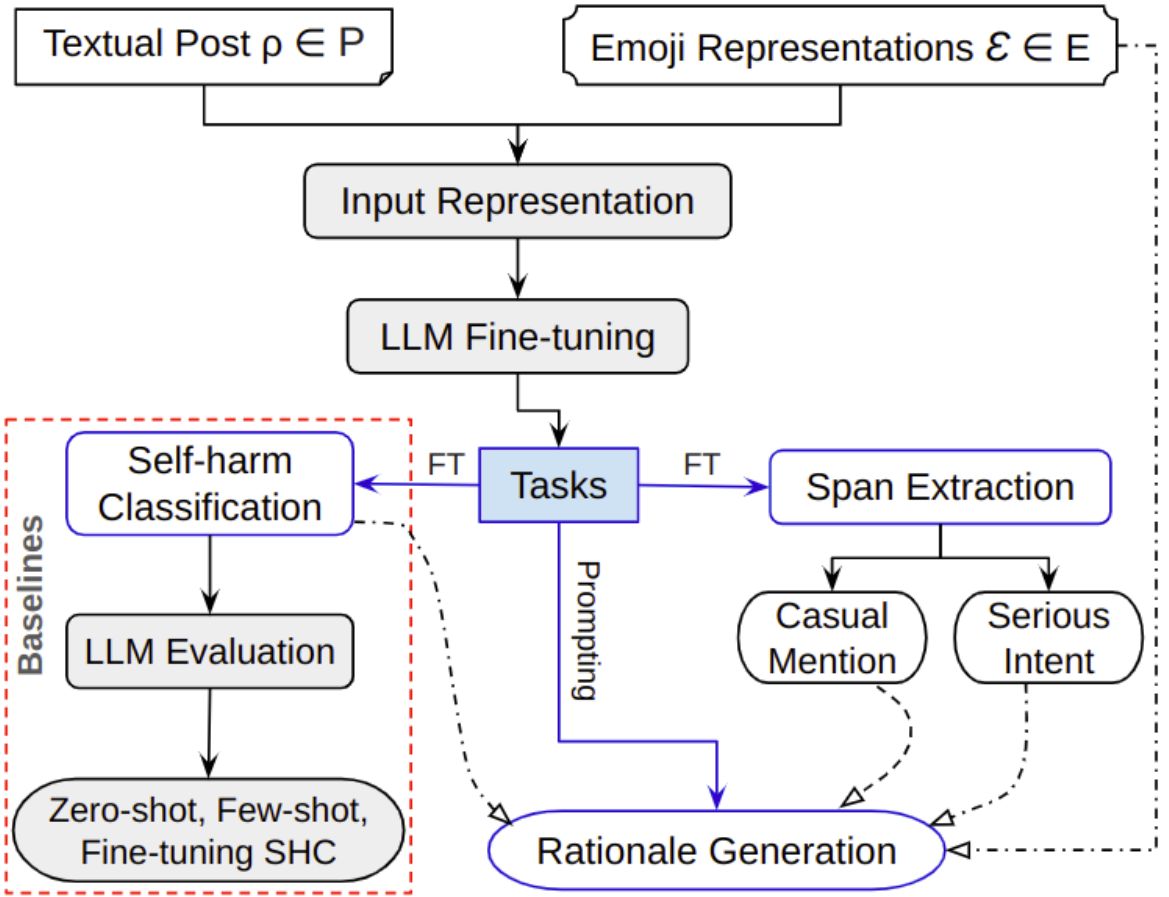}}
\caption{LLM Fine-Tuning for Self-Harm Detection and Rationale Generation. The red dotted line encloses the baseline evaluation setup for self-harm detection.}\label{nsa_arch}
\end{figure}

The proposed framework employs a multitask fine-tuning paradigm designed to integrate seamlessly with an LLM jointly addressing self-harm classification, span extraction for casual mentions (CM) and serious intents (SI), and rationale generation. Figure~\ref{nsa_arch} outlines the proposed approach, detailed in the following steps:


\textbf{Unified Input Representation}:
Table~\ref{tab:Aprompt} describes the prompt for fine-tuning our model which essentially integrates information from diverse sources (e.g., input post (text) and emoji interpretation (CESM-100)) into a single training instance without necessarily concatenating them. 

\textbf{Task-Specific Outputs with Shared Representations:}
The framework relies on the encoder-decoder architecture of the chosen LLM (e.g., Llama or Alpaca) to produce task-specific outputs:
\begin{itemize}[nolistsep]
    \item \textbf{Self-Harm Classification}: Predict a binary label $y \in {0, 1}$ indicating self-harm or non self-harm.
    \item \textbf{Span Extraction}: Predict CM spans $S_{CM} \subset \rho$ and SI spans $S_{SI} \subset \rho$ using sequence tagging layers aligned with the input tokens.
\end{itemize}

\begin{itemize}[nolistsep]
    \item \textbf{Rationale Generation}: The LLM is further prompted to generate rationales that explain the model’s decisions, explicitly referencing CM and SI spans as well as CESM-100-based emoji interpretations.
    
\end{itemize}

\textbf{Prompt-Driven Fine-Tuning:}
The framework utilizes task-specific prompts (Table~\ref{tab:Aprompt} in the Appendix) to adapt the LLM for multitask learning. These prompts guide the model to address each task cohesively while leveraging shared input representations. The prompts remain flexible and can be tailored to the specific capabilities of the LLM.

\textbf{Integrated Optimization:}
A multitask loss function ensures balanced learning across all tasks during fine-tuning. We utilized binary cross-entropy loss for the self-harm classification task and sparse categorical cross-entropy loss for the causal mention and serious intent extraction tasks. Specifically, the sparse categorical cross-entropy loss was applied to optimize the prediction of the start and end positions of causal spans by calculating the cross-entropy between the true indices and the predicted probability distributions across all tokens in the sequence.


\begin{table*}[!ht]
\centering
\caption{Performance Comparison of Llama-3.1-8B-Instruct, Mental Alpaca, and MentaLLaMA-chat-7B models across various tasks in different experimental settings. \textit{Abbreviations:} SHC (Self-harm Classification), CMSE (Casual Mention Span Extraction), SISE (Serious Intent Span Extraction), RG (Rationale Generation). The values in parentheses represent the variance for each score.}
\label{tab:perf}
\renewcommand{\arraystretch}{1.28}
\setlength\tabcolsep{8pt}
\begin{adjustbox}{max width=1\textwidth}
\begin{tabular}{c|c|cc|cccc}
\hline
\textbf{Model} & \textbf{SHC} & \textbf{CMSE} & \textbf{SISE} & \multicolumn{4}{|c}{\textbf{Rationale Generation (RG)}} \\
& \textbf{F1} & \textbf{F1} & \textbf{F1} & \textbf{Relevance} & \textbf{Coherence} & \textbf{Readability} & \textbf{SemSim} \\ \hline
\multicolumn{7}{c}{\textbf{Zero-shot Prompting for SHC and RG}}\\ \hline
Llama 3 & 0.74 (±0.002) & - & - & 0.75 (±0.003) & 0.76 (±0.001) & 0.74 (±0.001) & 0.75 (±0.003) \\
Mental-Alpaca & 0.76 (±0.003) & - & - & 0.73 (±0.004) & 0.75 (±0.003) & 0.74 (±0.002) & 0.74 (±0.001) \\
MentalLlama & 0.72 (±0.001) & - & - & 0.71 (±0.003) & 0.72 (±0.001) & 0.71 (±0.002) & 0.72 (±0.004) \\
\hline
\multicolumn{7}{c}{\textbf{Few-shot Prompting for SHC and RG}}\\ \hline
Llama 3 & 0.79 (±0.002) & - & - & 0.80 (±0.001) & 0.81 (±0.004) & 0.79 (±0.003) & 0.80 (±0.001) \\
Mental-Alpaca & 0.80 (±0.004) & - & - & 0.78 (±0.001) & 0.79 (±0.004) & 0.78 (±0.002) & 0.79 (±0.003) \\
MentalLlama & 0.78 (±0.003) & - & - & 0.77 (±0.003) & 0.76 (±0.001) & 0.77 (±0.001) & 0.78 (±0.002) \\
\hline
\multicolumn{7}{c}{\textbf{Fine-tuning (for SHC) + Prompting (for RG)}}\\ \hline
Llama 3 & 0.83 (±0.003) & - & - & 0.85 (±0.003) & 0.85 (±0.004) & 0.84 (±0.003) & 0.84 (±0.001) \\
Mental-Alpaca & 0.82 (±0.002) & - & - & 0.83 (±0.004) & 0.83 (±0.001) & 0.83 (±0.002) & 0.83 (±0.003) \\
MentalLlama & 0.81 (±0.002) & - & - & 0.81 (±0.003) & 0.82 (±0.002) & 0.80 (±0.001) & 0.81 (±0.002) \\
\hline \hline
\multicolumn{7}{c}{\textbf{\textit{[Ours]} Multitask (MT) Fine-tuning (for SHC, CMSE, and SISE) + Prompting (for RG) + w/ CESM-100}}\\ \hline
Llama 3 & 0.88 (±0.002) & 0.85 (±0.003) & 0.84 (±0.003) & 0.89 (±0.003) & 0.89 (±0.001) & 0.88 (±0.003) & 0.88 (±0.001) \\
Mental-Alpaca & 0.86 (±0.001) & 0.83 (±0.002) & 0.82 (±0.003) & 0.87 (±0.003) & 0.87 (±0.003) & 0.87 (±0.001) & 0.86 (±0.002) \\
MentalLlama & 0.85 (±0.001) & 0.80 (±0.004) & 0.81 (±0.003) & 0.85 (±0.003) & 0.85 (±0.001) & 0.85 (±0.003) & 0.85 (±0.001) \\
\hline
\multicolumn{7}{c}{\textbf{\textit{[Ablation 1]} MT Fine-tuning (for SHC, CMSE, and SISE) + Prompting (for RG) + w/o CESM-100}}\\ \hline
Llama 3 & 0.84 (±0.001) & 0.83 (±0.003) & 0.81 (±0.004) & 0.86 (±0.003) & 0.85 (±0.003) & 0.85 (±0.002) & 0.85 (±0.001) \\
Mental-Alpaca & 0.82 (±0.003) & 0.81 (±0.001) & 0.80 (±0.003) & 0.84 (±0.003) & 0.84 (±0.003) & 0.84 (±0.002) & 0.83 (±0.004) \\
MentalLlama & 0.83 (±0.002) & 0.78 (±0.001) & 0.79 (±0.003) & 0.83 (±0.003) & 0.82 (±0.003) & 0.83 (±0.001) & 0.82 (±0.003) \\
\hline
\multicolumn{7}{c}{\textbf{\textit{[Ablation 2]} Fine-tuning (for SHC) + Prompting (for RG) + w/ CESM-100}}\\\hline
Llama 3 & 0.86 (±0.001) & - & - & 0.86 (±0.004) & 0.86 (±0.003) & 0.85 (±0.002) & 0.85 (±0.001) \\
Mental-Alpaca & 0.83 (±0.002) & - & - & 0.84 (±0.001) & 0.85 (±0.003) & 0.84 (±0.003) & 0.84 (±0.004) \\
MentalLlama & 0.82 (±0.003) & - & - & 0.83 (±0.003) & 0.83 (±0.001) & 0.82 (±0.003) & 0.82 (±0.001) \\
\hline
\end{tabular}
\end{adjustbox}
\end{table*}

\section{Experimental Setup}
To ensure thorough evaluation, we carefully chose models relevant to self-harm detection, balancing model size with hardware limitations (NVIDIA K80 GPU, 24 GB GDDR5 memory) to enable comprehensive fine-tuning. Our dataset is split 80/20 for training and testing, with synthetic posts used only in training. We report results from the average of five experimental runs.

We report the F1 score for the self-harm classification task. For the span extraction tasks, we follow recent works \cite{poria2021recognizing,ghosh2022multitask} and use F1 to evaluate the quality of the extracted spans. To effectively evaluate the quality of generated rationales for self-harm detection, we employ a combination of Relevance \cite{teh2024impact}, Coherence \cite{teh2024impact}, Readability \cite{flesch2007flesch}, and Semantic Similarity \cite{faysse2023revisiting} measures. These metrics ensure that the generated rationales are not only accurate and relevant but also comprehensible and logically consistent. Detailed definitions, implementation specifics, and hyper-parameters are provided in the appendix.

Motivated by recent studies \cite{sarkar2023testing,behnamghader2024llm2vec,dukic2024looking} and prior experience, we opted for decoder-only models over encoder only or encoder-decoder models for self-harm classification and span extraction, as they excel in tasks requiring robust contextual understanding and sequence labeling. In this context, we chose one open-domain LLM (Llama-3.1-8B-Instruct \cite{llama3modelcard}) and two domain-specific LLMs (MentaLLaMA-chat-7B \cite{yang2024mentallama}, Mental Alpaca \cite{xu2024mental}) for our evaluation of SHINES dataset and effectiveness of CESM-100. Model details can be found in section~\ref{modeldetails} of the appendix.

This study focuses on evaluating open-source pre-trained LLMs of comparable sizes, including both open-domain and domain-specific models. Although larger proprietary models like GPT-3.5 Turbo or o1-mini offer potential advantages, our choices were guided by hardware availability. Within these constraints, we optimized resources to ensure a fair and consistent setup. Despite limitations, the study provides meaningful insights into open-access model performance and motivates future evaluations with larger models.
 
We compare multiple LLMs across zero-shot, few-shot, and fine-tuning setups to ensure a robust evaluation. In the zero-shot setting, the LLM receives a textual post and task description without prior examples or fine-tuning, relying solely on pre-trained knowledge. The few-shot setting introduces 2 or 5 examples to help the LLM better recognize self-harm-related content and generate logical rationales. In the fine-tuning setup, we train the LLMs on SHINES for self-harm classification and prompt them to generate rationales supporting their classification decisions. For prompt details, see Table~\ref{tab:prompts} in Appendix~\ref{app:prompts}.

\section{Results and Discussion}
Table~\ref{tab:perf} presents performance trends of Llama-3.1-8B-Instruct, Mental-Alpaca-7B, and MentalLlama-7B across self-harm classification (SHC), span extraction (CMSE and SISE), and rationale generation (RG) tasks.

Few-shot prompting improves performance over zero-shot but remains inferior to fine-tuning. For example, Mental-Alpaca’s SHC F1 score increases from 0.76 (zero-shot) to 0.80 (few-shot) and further to 0.82 with fine-tuning. Multitask fine-tuning surpasses other strategies by leveraging shared knowledge, with Llama 3 achieving an SHC F1 of 0.88 when incorporating CESM-100, compared to 0.83 without it in single-task fine-tuning. Similarly, multitask fine-tuning enhances CMSE and SISE performance, with Llama 3 attaining the best F1 scores of 0.85 and 0.84, respectively.

Rationale generation benefits from improved upstream performance, with Llama 3 achieving a semantic similarity score of 0.88 in generated rationales, compared to 0.85 without CESM-100. The quality of rationales—measured by relevance, coherence, and readability—improves with stronger contextual reasoning from SHC and span extraction outputs. RG serves as a crucial post-task component, focusing on explainability (why the model predicted self-harm vs. non-self-harm) rather than interpretability (how it arrived at the decision).

We conducted statistical evaluations comparing SHC with CESM-100 to its ablated counterpart (without CESM-100) across Llama, Mental-Alpaca, and Mental-Llama configurations. Paired t-tests \cite{dietterich1998approximate} were chosen over Hoeffding inequality \cite{hoeffding1994probability} because the aggregated F1 scores are derived from non-i.i.d. data, making paired t-tests more suitable for capturing dependent relationships across runs. Results from 10 runs confirmed statistically significant F1 improvements, with p-values below 0.05 (e.g., Llama: p = 0.0198, Mental-Alpaca: p = 0.017, Mental-Llama: p = 0.009). For other setups, our proposed multitask model demonstrated substantial improvements: +12\% F1 over zero-shot, +7\% F1 over few-shot, and +4\% F1 over fine-tuned setups (averaged across models). Given these large margins, significance testing was not initially performed.

\begin{table*}[!ht]
\centering
\caption{Predictions of the models for different posts. Blue: correct predictions; red: wrong predictions.}
\label{table:predictions}
\begin{adjustbox}{max width=1\textwidth}
\renewcommand{\arraystretch}{1.28}
\setlength\tabcolsep{5pt}
\begin{tabular}{p{4.7cm}|p{12.7cm}}
\hline
\textbf{Post Text} & \textbf{Predictions and Rationales} \\
\hline
\small \textbf{Post 1:} ``I've been struggling with self-harm for a while now. Sometimes it feels like the only way to release the pain inside me. \includegraphics[scale=0.05]{images/pensive.png}\includegraphics[scale=0.05]{images/broken_heart.png}\includegraphics[scale=0.05]{images/knife.png} It's like a dark cloud following me everywhere I go, \includegraphics[scale=0.05]{images/new_moon.png} but in those moments, the physical pain seems to overshadow the emotional turmoil. \includegraphics[scale=0.05]{images/rain_cloud.png}\includegraphics[scale=0.05]{images/sparkles.png} I know it's not a healthy coping mechanism, but it's hard to resist the urge when everything feels too much to handle. \includegraphics[scale=0.12]{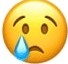}\includegraphics[scale=0.05]{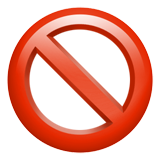} I wish I could find a different outlet, a way to express my emotions without causing myself harm. \includegraphics[scale=0.05]{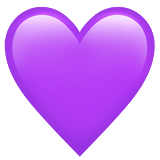}\includegraphics[scale=0.05]{images/muscle.png}" & 

\textbf{Llama 3:} \textcolor{blue}{Self-Harm}

\small \textit{Rationale:} The user openly discusses their struggle with self-harm as a means to alleviate inner pain, emphasizing the severity of their emotional distress. Key spans, such as ``Sometimes it feels like the only way to release the pain inside me", ``It's like a dark cloud following me everywhere I go", and ``It's hard to resist the urge when everything feels too much to handle", highlight significant distress and contemplation of self-harm. The presence of emojis like \includegraphics[scale=0.05]{images/knife.png} (which directly signifies self-harm), \includegraphics[scale=0.05]{images/broken_heart.png} (indicating intense emotional pain), and \includegraphics[scale=0.12]{images/low_sad.jpg} (reflecting deep sadness) reinforces the classification. 

\textbf{Mental-Alpaca:} \textcolor{blue}{Self-Harm}

\textit{Rationale:} Post explicitly describes self-harm as a coping mechanism for severe emotional pain. User expresses strong urge to self-harm despite desire for healthier coping strategies.

\textbf{MentalLlama:} \textcolor{blue}{Self-Harm}

\textit{Rationale:} The post is classified as self-harm due to direct references to self-harm and emotional distress. Phrases like ``the only way to release the pain" and emojis, such as self-harm and  emotional pain indicate significant self-harm behavior, despite the user’s desire for healthier coping methods. \\

\hline
\small \textbf{Post 2: }``Struggling with self-harm again. It's like a never-ending cycle \includegraphics[scale=0.05]{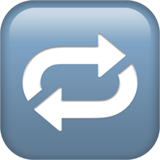}. The scars on my skin tell a story I wish I could erase. Sometimes the pain is the only thing that feels real \includegraphics[scale=0.05]{images/broken_heart.png}. Trying to hold on to hope, but it feels like I'm drowning in darkness \includegraphics[scale=0.05]{images/new_moon.png}. \#mentalhealthmatters \#recoveryjourney \includegraphics[scale=0.05]{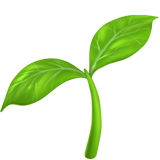}\includegraphics[scale=0.05]{images/muscle.png}\includegraphics[scale=0.05]{images/pensive.png}" & 

\small \textbf{Mental-Alpaca:} \textcolor{red}{Non Self-Harm}

\textit{Rationale:} The post describes emotional challenges without explicit references to self-harm. While acknowledging past struggles, the content focuses on recovery and personal growth. The use of hopeful and empowering language suggests a positive outlook and a desire for support, rather than active self-harm.

\textbf{MentalLlama:} \textcolor{red}{Non Self-Harm}

\textit{Rationale:} The post is classified as non-self-harm because it discusses emotional struggle without explicit references to self-harm. Phrases like ``Struggling with self-harm again" and emojis such as growth and strength indicate a focus on recovery and resilience, not self-harm. The content emphasizes hope and seeking support rather than self-harm behavior.

\textbf{Llama 3:} \textcolor{blue}{Self-Harm}

\textit{Rationale:} User expresses ongoing struggle with self-harm and describes it as a temporary relief from overwhelming pain. Despite hopes for recovery, the overall tone suggests a crisis situation.\\

\hline
\end{tabular}
\end{adjustbox}
\end{table*}

\subsection{Ablation Study}
\textbf{Impact of CESM-100 Integration:}
The ablation study shows that CESM-100 has a measurable impact on performance by providing contextual examples that improve the models' ability to capture subtle distinctions. When excluded, performance drops across all tasks. For instance: Mental-Alpaca's CMSE F1 score decreases from 0.83 (with CESM-100) to 0.81 (without CESM-100). Mental-Alpaca's SISE F1 score drops from 0.82 to 0.80. For SHC, integrating CESM-100 raises Llama 3's F1 score from 0.84 (without CESM-100) to 0.88 in multitask fine-tuning. For CMSE, CESM-100 enables MentalLlama to achieve an F1 score of 0.80, outperforming its performance without CESM-100 (0.78). These findings highlight CESM-100's ability to supply key contextual signals that improve the model's reasoning about nuanced spans and improve generalization performance.

\textbf{Impact of Multitask Fine-tuning:}
The ablation study (Ablation 2) highlights the advantages of multitask fine-tuning. Fine-tuning SHC in combination with CMSE and SISE tasks leads to strong performance outcomes. For instance, Llama 3’s F1 score for SHC saw a notable 3-point drop when shifted from multitask fine-tuning to single-task fine-tuning, while MentalLlama and MentalAlpaca exhibited decreases of 2 points each under similar conditions. Multitask fine-tuning improves performance by enabling shared knowledge across tasks, fostering better generalization, and enhancing downstream performance in both classification and rationale generation tasks.

\textbf{Impact of Emoji by Adding Noise:}
To further investigate the impact of emojis on the performance of LLMs within our framework, we introduced noise into the training data in two main ways: i) altering the position of emojis within the training posts, and ii) replacing emojis in a post with randomly selected ones from CESM-100, which may or may not be contextually relevant. These manipulations were applied to 20\% of the training posts containing emojis. The observed 2.5\% drop in self-harm classification F1 score is substantial, given the small dataset size and the skewed distribution of emojis between self-harm and non-self-harm posts. This observation underscores the critical role of emojis in enabling effective contextual understanding by LLMs during training, highlighting the need for further exploration to uncover insights.

\subsection{Qualitative analysis}
We performed a thorough analysis of the predictions across the three models revealing their varying sensitivities to self-harm indicators and context. Table~\ref{table:predictions} reports some sample test instances from the \textit{SHINES} test set with the observed model ouputs. While most models align in their overall assessments, discrepancies arise particularly with nuanced language and emotional tone.

\begin{itemize}[nolistsep]
\item \textit{Post 1:} This post directly discusses ongoing self-harm struggles. All models (Llama 3, Mental-Alpaca, and MentalLlama) classified this post as self-harm. They accurately identified it as self-harm due to explicit mentions and emojis related to self-harm and emotional distress, despite the user's acknowledgment of the harmful nature of their coping mechanism.
\item \textit{Post 2:} This post discusses persistent self-harm and emotional pain. Models varied in classification, with Llama 3 labeling it as self-harm, while Mental-Alpaca and MentalLlama categorized it as non self-harm. Models emphasizing explicit self-harm references and emojis performed better. Contextual elements like hashtags and emotive language may cause some models to underweight serious self-harm indicators.
\end{itemize}

Llama, being the best-performing model among our experimental LLMs, we further performed a qualitative analysis of Llama's performance with and without our framework. Our qualitative analysis revealed that prompt-based fine-tuning and emoji awareness from CESM-100 significantly enhanced Llama's reasoning ability, leading to improved decision-making capability compared to its out-of-the-box performance. For intance, Llama classified the following example as self-harm, however, with our prompt-based fine-tuning, incorporating emoji awareness from CESM-100, Llama correctly classified it as non-self-harm.

\textit{"My inner garden was overgrown with weeds of worry and thorns of fear \includegraphics[scale=0.05]{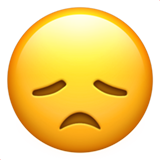}. I tried to prune them back, to cultivate the good, but the thorns kept pricking me, drawing blood. It was a constant battle, a struggle to keep the beautiful flowers from being choked \includegraphics[scale=0.05]{images/wilted_flower.png} out by the invasive weeds."}

In interest of space, we moved the error analysis discussion to Section~\ref{App:errana} of the appendix.

\section{Conclusion}
This study tackles self-harm detection on social media by integrating advanced LLMs with contextual emoji analysis in a multi-task learning framework. We introduce the Centennial Emoji Sensitivity Matrix (CESM-100) and SHINES dataset, which provide nuanced emoji interpretations and a comprehensive annotated corpus for self-harm classification, CM and SI span detection, and rationale generation. Our multi-task learning approach improves classification accuracy, span detection, and rationale generation. CESM-100 enhances semantic similarity scores and contextual reasoning in LLMs, highlighting the value of domain-specific resources for mental health detection. These findings support expanding datasets across diverse contexts.

Future work includes leveraging temporal behavioral patterns, user network interactions, and profile metadata to improve accuracy. Enhancing model explainability and addressing biases and privacy concerns will be crucial for ethical deployment.

\section*{Ethical Considerations and Limitations}
The introduced dataset contains social media posts discussing self-harm, including explicit and implicit mentions of distress, emotional struggles, and related themes. Some content may be triggering or distressing to certain individuals. Reader discretion is advised. If you are experiencing distress or require support, please consider reaching out to a mental health professional or a crisis support service in your region.

The study is not intended for clinical application in its current form. Its primary objective is to investigate how computational methods can be improved to better analyze social media discourse. While our findings contribute to the understanding of self-harm detection, they should be interpreted within the scope of computational research rather than as a diagnostic or intervention tool.

\begin{itemize}[nolistsep]

\item \textit{Non-Clinical Use:} The predictions and rationales generated by large language models (LLMs) in this study are intended strictly for research purposes. If made publicly accessible without proper safeguards, they could be misinterpreted by non-experts, leading to incorrect self-diagnosis or harmful actions. These models are not substitutes for professional psychiatric advice, and we strongly emphasize the importance of consulting qualified mental health professionals for self-harm concerns.

\item \textit{Bias and Misuse:} LLMs inherently reflect biases present in their training data, and our framework is not immune to this challenge. These biases can subtly influence predictions and rationales, particularly in the sensitive domain of self-harm detection. The risk of misinterpretation or amplification of stereotypes necessitates rigorous scrutiny and mitigation strategies to ensure ethical use. Careful selection and evaluation of training data, alongside diverse representation in future datasets, will help address these issues.

\item \textit{Platform Bias from Reddit Data:}
Our reliance on a single platform for data collection may impact the generalizability of our findings. Due to the sensitive nature of our research topic and the stigma surrounding self-harm discussions, identifying suitable online platforms was challenging. After rigorous investigation, Reddit emerged as the most viable platform for developing a relevant corpus. To mitigate potential bias, we employed a comprehensive approach, incorporating data from a wide range of subreddits related to self-harm, as detailed in Appendix A.1. Furthermore, the selection of self-harm-related connotations was grounded in prior research and carefully chosen following an extensive background study. Platform bias cannot be entirely eliminated and we recognize this as a limitation and plan to expand our work to include data from diverse platforms as they become available and suitable for our objectives. 

\item \textit{Dataset Sensitivity:} The \textit{SHINES} dataset contains posts that may include distressing content or references to self-harm, which could provoke negative emotions in developers or users. This necessitates cautious use, complemented by access to support resources for individuals working with the data. Furthermore, while the dataset is invaluable for advancing self-harm detection, its use must be accompanied by appropriate disclaimers to prevent unintended consequences or misuse.

\item \textit{Cultural and Contextual Variability:} While the \textit{CESM-100} and \textit{SHINES} datasets are designed to address cultural nuances in self-harm language and emoji use, cultural variability remains a significant challenge. \textit{CESM-100} integrates emoji interpretations spanning the timeline 2010-2024 and leverages insights from Emojipedia to reduce temporal and cultural biases. However, emojis and language evolve, and interpretations may vary across contexts. Future research must ensure these datasets remain adaptive, incorporating broader cultural insights and additional context-specific variations.

We recognize that platform-specific communication norms, including the sarcastic or non-literal use of emojis, may differ significantly across online communities. While our distributional analysis suggests that sarcastic emoji usage is rare within our dataset, there is possibility that other platforms might exhibit different trends. Addressing this limitation is an important direction for future research.

\item \textit{Clinical Validation and Relevance:} Both \textit{SHINES} and \textit{CESM-100} are informed by a clinical perspective, with their annotation schema and emoji interpretations validated by a psychiatrist with over 19 years of expertise in mental health. However, while these resources offer reliability and adaptability, they are not a substitute for clinical datasets collected directly from patient interactions. Future collaborations with clinical experts and institutions are essential to expand the applicability and validity of these resources.

\item \textit{Generalizability and Scalability:} Although \textit{SHINES} and \textit{CESM-100} address modern social media trends, their applicability across platforms, demographics, and linguistic variations may be limited. Expanding these datasets to other platforms and languages, along with real-time adaptability, remains a priority for future work. This road map is crucial for capturing the evolving dynamics of self-harm expressions and ensuring cross-cultural reliability.
\end{itemize}

\section*{Use of AI Assistants} ChatGPT was used to enhance the presentation quality of the text in this paper, including proofreading and detecting any typos or other errors.
\bibliography{custom}

\appendix

\section{Appendix}

\subsection{Sub-reddits Used for Data Collection}\label{a.1}
Below is the full list of sub-reddits considered for posts collection to build our SHINES dataset.

\texttt{`mentalhealth', `traumatoolbox', `TrueOffMyChest', `anxiety', `BPD', `depression', `suicidewatch', `mentalillness', `selfharm', `offmychest', `vent', `suicidalthinking', `anxiety', `operation', `stress', `competition', `workPressure', `sports', `heavyHeart', `mentalhealth', `mentalillness', `depression', `politics', `askatherapist', `socialskills', `BodyAcceptance', `bodyneutrality', `BodyNeutrality', `Mindfulness', `BipolarReddit', `ADHD', `bipolar', `positivity', `suicidewatch', `suicidalthinking', `officePolitics', `parenting', `selfinjurysupport', `medication', `characters', `nostalgia', `environment', `instagram', `relationships', `panera', `religion', `selfhelp'}.

\subsection{Data Filtering}\label{a:df}
We initially collected over 5,000 posts from relevant subreddits using API calls and keyword searches. To ensure data quality and relevance, we applied the following filtering criteria:
\begin{itemize}[nolistsep]
    \item \textit{Extremely Short Posts:} Posts with fewer than 3–5 words were excluded as non-informative.
\item \textit{Title-Only Posts:} Posts lacking meaningful body content were removed due to insufficient context for analysis.
\item \textit{Duplicate Posts:} Copy-pasted or repeated posts by the same user across subreddits were filtered out to avoid redundancy.
\item \textit{Noisy Posts:} Posts with excessive special characters, improper formatting, or non-English content were excluded.
\item \textit{Non-Textual Posts:} Posts consisting solely of links or images without substantial text were omitted, as the focus was on text analysis.
\end{itemize}

\subsection{Guidelines for Generating Emoji Interpretations Relevant to Self-Harm for CESM-100}\label{guidelines}

When interpreting emojis in the context of self-harm, the following guidelines were followed to ensure accurate and sensitive representation:

\begin{table*}[!ht]
\centering
\caption{Emoji Usage Examples and Contextual Phrases. CM: Casual Mention, SI: Serious Intent.}
\label{tab:emoji_examples}

\begin{tabular}{c|c|p{9cm}}
\hline
Emoji & Category & Contextual Phrase \\
\hline
\includegraphics[scale=0.05]{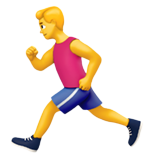}\includegraphics[scale=0.05]{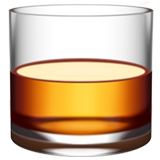}\includegraphics[scale=0.05]{images/knife.png}\includegraphics[scale=0.05]{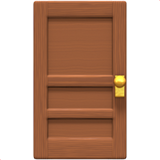} & Metaphorical Representation & Some people run away, others drink, and then there are those of us who only find escape through the pain. \includegraphics[scale=0.05]{images/runner.png}\includegraphics[scale=0.05]{images/tumbler_glass.png}\includegraphics[scale=0.05]{images/knife.png}\includegraphics[scale=0.05]{images/door.png}\\
\hline
\includegraphics[scale=0.007]{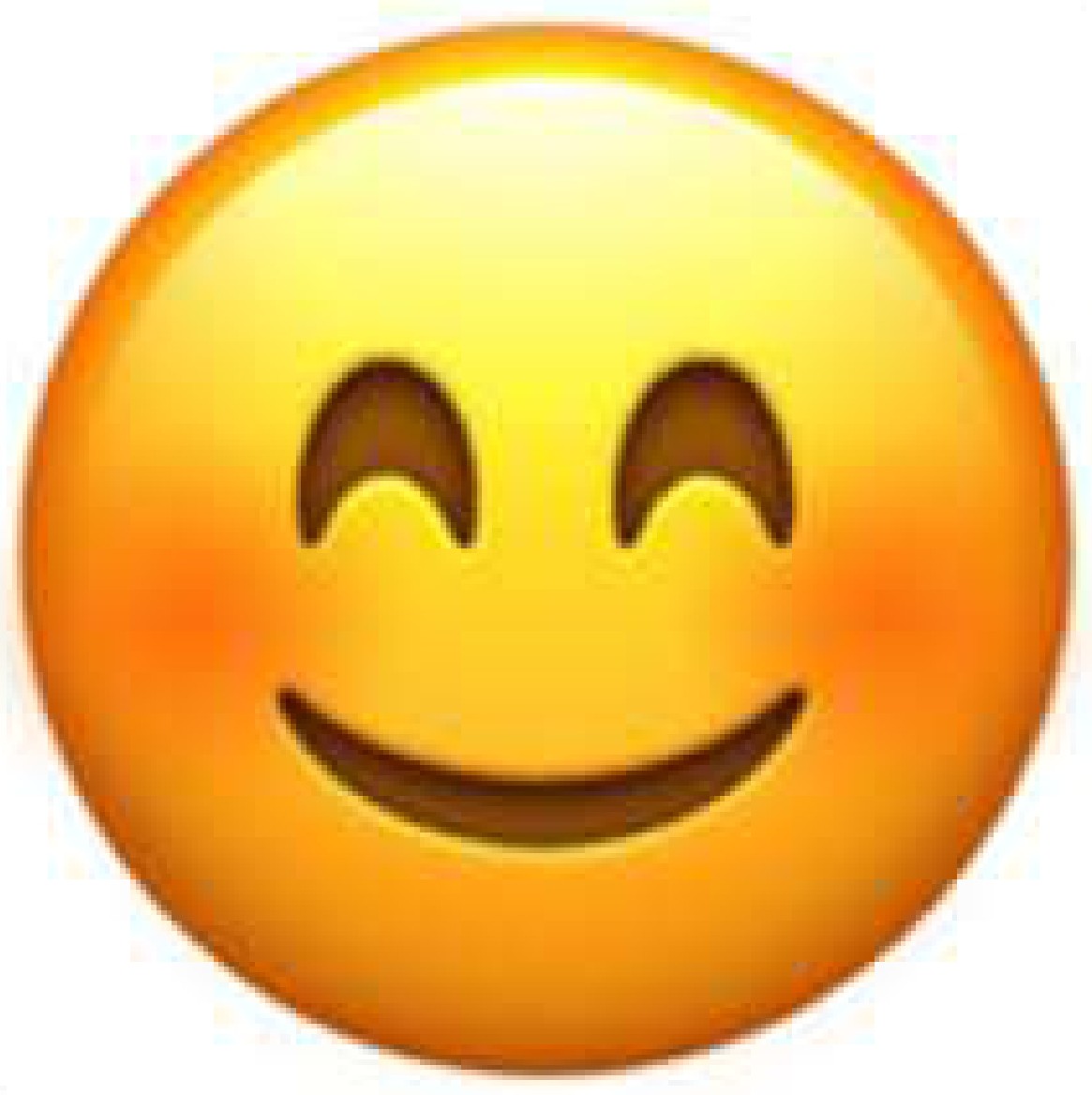}\includegraphics[scale=0.05]{images/broken_heart.png} & Direct Representation & Smiling outside, breaking inside. \includegraphics[scale=0.007]{images/Happy_smile.jpg}\includegraphics[scale=0.05]{images/broken_heart.png}\\ \hline
\includegraphics[scale=0.05]{images/knife.png}\includegraphics[scale=0.007]{images/drop-of-blood.653x1024.png}\includegraphics[scale=0.12]{images/black_heart.png}\includegraphics[scale=0.06]{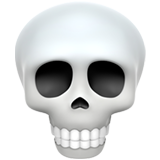} & Semantic List & The thoughts keep coming back. I don’t know how much longer I can resist. \includegraphics[scale=0.05]{images/knife.png}\includegraphics[scale=0.007]{images/drop-of-blood.653x1024.png}\includegraphics[scale=0.12]{images/black_heart.png}\includegraphics[scale=0.06]{images/skull.png}\\
\hline
\end{tabular}

\end{table*}

\begin{enumerate}[label=\textbf{\arabic*.},nolistsep]
    \item \textbf{Contextual Sensitivity}  
    Interpret emojis considering both their usual meanings and the emotional context of self-harm.  
    Focus on how the emoji might reflect feelings, actions, or intentions commonly associated with self-harm or emotional distress.

    \item \textbf{Categorization by Intent}  
   Categorize emoji usage into \textit{Casual Mention} or \textit{Serious Intent} based on:  
    \begin{itemize}[nolistsep]
        \item \textit{Casual Mention:} Reflects indirect or non-serious emotional expressions, often used metaphorically or humorously. 
        \item \textit{Serious Intent:} Indicates direct or severe expressions of emotional pain or potential self-harm intentions.
    \end{itemize}

    \item \textbf{Usage Frequency}  
    Assess the frequency of occurrence for each category:
    \begin{itemize}[nolistsep]
        \item \textit{Low:} Rarely used but significant when present.
        \item \textit{Medium:} Moderately common, with consistent thematic relevance. 
        \item \textit{High:} Frequently used, either to express emotions or mask them.
    \end{itemize}

    \item \textbf{Emotional Nuances}  
 Account for both explicit and implicit meanings: 
    \begin{itemize}[nolistsep]
        \item \textit{Explicit:} Clearly conveys self-harm or distress (e.g., \includegraphics[scale=0.13]{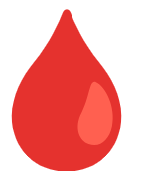} Drop of Blood, \includegraphics[scale=0.05]{images/knife.png} Kitchen Knife). 
        \item \textit{Implicit:} Suggests emotional pain, coping, or hope (e.g., \includegraphics[scale=0.05]{images/seedling.png}, \includegraphics[scale=0.05]{images/sparkles.png}).
    \end{itemize}

    \item \textbf{Dual Nature of Emojis}  
Recognize that some emojis have dual meanings: 
    \begin{itemize}[nolistsep]
        \item \textit{Positive masking negative:} Emojis like \includegraphics[scale=0.05]{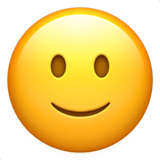} Smiling Face might hide deeper emotional struggles. 
        \item \textit{Negative with hopeful undertones:} Emojis like \includegraphics[scale=0.15]{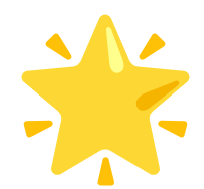} Glowing Star symbolize hope amidst emotional pain.
    \end{itemize}

    \item \textbf{Alignment with Context}  
Interpret emojis based on their role within the specific narrative or post:  
    \begin{itemize}[nolistsep]
        \item Ensure that the interpretation aligns with the surrounding text or implied emotional state.        \item Avoid overgeneralization or detachment from the user's intended meaning.
    \end{itemize}

    \item \textbf{Visual and Emotional Symbols}  
  Emojis with visual metaphors (e.g.,  Dove for peace \includegraphics[scale=0.18]{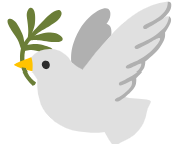},  Black Heart \includegraphics[scale=0.15]{images/black_heart.png} for depressive emotions) should be mapped to their emotional significance in the self-harm context.

    \item \textbf{Ethical Sensitivity}  
Maintain ethical responsibility by:     
\begin{itemize}[nolistsep]
        \item Avoiding stigmatizing interpretations.  
        \item Acknowledging that not all uses of emojis are linked to self-harm but could represent broader emotional expressions.
    \end{itemize}

    \item \textbf{Validation and Refinement}  
 Ensure interpretations are:  
\begin{itemize}[nolistsep]
        \item Validated by mental health professionals to maintain accuracy and reduce bias. 
        \item Iteratively refined based on feedback from stakeholders (e.g., annotators, psychologists).
    \end{itemize}
\end{enumerate}

\subsection{Emoji Usage in Self-Harm Expressions}\label{App:Emojianalysis}

We performed a qualitative analysis of emoji usage within self-harm expressions, incorporating relevant examples from the CESM-100 matrix. Our observations are as follows:

\paragraph{Emoji Choices and Composition}  
Specific emojis convey different emotional tones. emojis like \includegraphics[scale=0.05]{images/skull.png} (Skull) and \includegraphics[scale=0.05]{images/coffin.png} (Coffin) are closely tied to death and serious intent, while \includegraphics[scale=0.05]{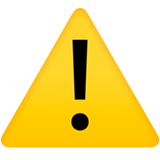} (Warning Sign) emphasizes danger or risk. Direct representations of self-harm, such as \includegraphics[scale=0.05]{images/knife.png}(Kitchen Knife) and \includegraphics[scale=0.02]{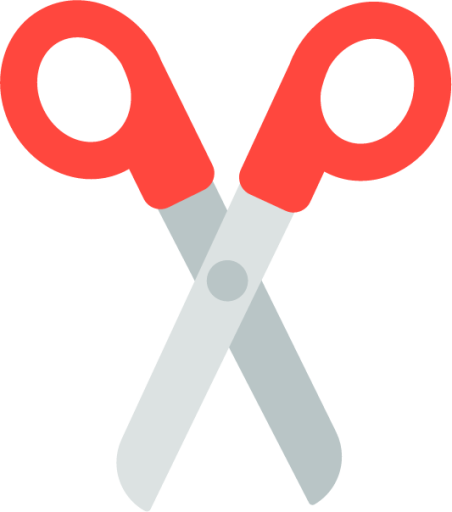}(Scissors), indicate contemplation of harmful actions. Interestingly, these violent and serious emojis often appear alongside supportive or neutral ones like \includegraphics[scale=0.05]{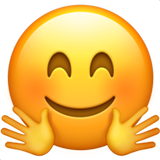} (Hugging Face) and \includegraphics[scale=0.05]{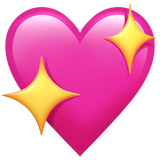}, reflecting internal conflict between despair and a desire for empathy. Emojis like \includegraphics[scale=0.05]{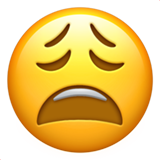} and \includegraphics[scale=0.05]{images/broken_heart.png} (Broken Heart) convey vulnerability and an urgent plea for support.

\paragraph{Interaction with Context}  
Emojis gain meaning from their textual surroundings. For instance, \textit{"I feel so alone \includegraphics[scale=0.05]{images/pensive.png}} emphasizes despair, while \includegraphics[scale=0.05]{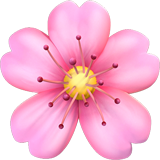} (Blossom) lightens the tone in \textit{"I'm having a rough day, but I’ll get through it \includegraphics[scale=0.05]{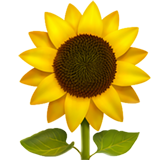}"} by introducing optimism.

\paragraph{Pragmatic and Contextual Uses}  
Emojis clarify emotional intent. For example, \textit{"I just wish I could escape \includegraphics[scale=0.05]{images/sob.png}..."} signals distress, while \textit{"Some days are tougher than others \includegraphics[scale=0.05]{images/sweat_smile.png}"} (Sweating Smile) conveys a more casual tone. Juxtaposing \includegraphics[scale=0.13]{images/low_sad.jpg} (crying face) and \includegraphics[scale=0.05]{images/broken_heart.png} (Broken Heart) signals a plea for empathy.

\paragraph{Strong Association of Certain Emojis with Self-Harm}  
\includegraphics[scale=0.05]{images/broken_heart.png} (Broken Heart), \includegraphics[scale=0.05]{images/muscle.png} (Flexed Biceps), \includegraphics[scale=0.05]{images/new_moon.png} (New Moon), \includegraphics[scale=0.05]{images/pensive.png}, and \includegraphics[scale=0.05]{images/knife.png} are the most frequently used emojis in self-harm posts.  
\includegraphics[scale=0.05]{images/broken_heart.png} appears 1337 times in self-harm contexts but only 56 times in non-self-harm contexts. \includegraphics[scale=0.05]{images/new_moon.png} appears exclusively in self-harm posts (1122 times), indicating its strong association with self-harm-related discourse. \includegraphics[scale=0.05]{images/knife.png} (Knife) appears 752 times in self-harm contexts and only 147 times in non-self-harm contexts, reinforcing its connection to distress.  

\paragraph{Support and Encouragement in Mental Health Conversations}  
\includegraphics[scale=0.05]{images/muscle.png} (Flexed Biceps) appears in 299 self-harm posts, suggesting that people use it to express resilience and inner strength while discussing self-harm.  
\includegraphics[scale=0.05]{images/purple_heart.png} (Purple Heart), \includegraphics[scale=0.17]{images/dove.png} (Dove), \includegraphics[scale=0.06]{images/rain_cloud.png} (Rain Cloud), and \includegraphics[scale=0.05]{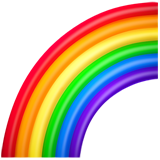} (Rainbow) are also heavily used in self-harm-related discussions, likely as symbols of hope, peace, and recovery.  
\includegraphics[scale=0.05]{images/hugging_face.png} (Hugging Face) appears 388 times in self-harm posts, showing its use in providing emotional support.  
\includegraphics[scale=0.05]{images/purple_heart.png} (Purple Heart) and \includegraphics[scale=0.17]{images/dove.png} are used exclusively in self-harm-related contexts, reinforcing their role in discussions about emotional pain and support.  

\paragraph{Usage of Specific Emojis in Self-Harm Contexts}  
Many emojis, such as \includegraphics[scale=0.05]{images/purple_heart.png} (Purple Heart), \includegraphics[scale=0.05]{images/new_moon.png} (New Moon), \includegraphics[scale=0.15]{images/dove.png} (Dove), and \includegraphics[scale=0.13]{images/black_heart.png} (Black Heart), appear only in self-harm posts, suggesting that these symbols are strongly linked to distress and mental health discussions. \includegraphics[scale=0.19]{images/lock.png} (Lock) appears almost exclusively in self-harm posts (449 times), possibly symbolizing feeling trapped or helpless.  

We also identified additional dimensions in emoji use within self-harm discussions, illustrated with CESM-100 examples:

\begin{itemize}[nolistsep]
    \item \textbf{Emotional Nuance:} Emojis like \includegraphics[scale=0.05]{images/sob.png} (Crying Face) enhance emotional depth in expressions of despair.
    \item \textbf{Supportive vs. Discomforting:} \includegraphics[scale=0.05]{images/hugging_face.png} (Hugging Face) fosters support, whereas \includegraphics[scale=0.05]{images/knife.png} (Kitchen Knife) underscores distress.
    \item \textbf{Coping Mechanisms:} Emojis such as \includegraphics[scale=0.05]{images/joy.png} (Face with Tears of Joy) appear in self-deprecating humor and coping strategies.
    \item \textbf{Community Engagement:} \includegraphics[scale=0.05]{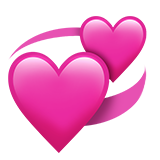} (Revolving Hearts) builds connection among users discussing mental health struggles.
    \item \textbf{Contrast in Self-Presentation:} \includegraphics[scale=0.05]{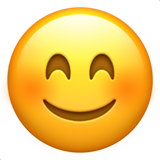} (Smiling Face with Smiling Eyes) and \includegraphics[scale=0.05]{images/pensive.png} (Pensive Face) illustrate emotional fluidity.
    \item \textbf{Intersection with Identity:} Emojis like \includegraphics[scale=0.05]{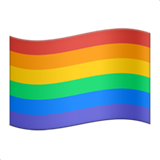} (Rainbow Flag) reflect personal identity and community experiences.
\end{itemize}

\subsection{Analysis of Serious Intent Emoji Combinations}

\paragraph{Dominance of Negative Emotions in Serious Intent Cases}  
In Serious Intent (SI) cases, certain emoji combinations such as \includegraphics[scale=0.05]{images/new_moon.png}\includegraphics[scale=0.05]{images/coffin.png}, \includegraphics[scale=0.05]{images/new_moon.png}\includegraphics[scale=0.05]{images/rain_cloud.png}, and \includegraphics[scale=0.05]{images/low_sad.jpg}\includegraphics[scale=0.05]{images/broken_heart.png}\includegraphics[scale=0.05]{images/knife.png} are among the most frequently occurring. These combinations often include emojis symbolizing darkness, storms, sadness, pain, and self-harm, reflecting intense emotional distress. Their recurring use underscores the severity of the emotions expressed.

\paragraph{High Occurrence of Nature and Weather-Related Symbols}  
Many Serious Intent cases also prominently feature weather-related emojis like \includegraphics[scale=0.05]{images/new_moon.png} (new moon), \includegraphics[scale=0.05]{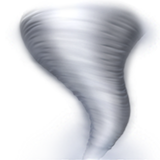} (tornado), and \includegraphics[scale=0.06]{images/rain_cloud.png} (rain). This trend suggests that individuals commonly use weather metaphors to communicate mental distress. Rain, storms, and darkness are particularly linked to feelings of depression and sadness, signaling the emotional weight carried by these symbols in the context of serious intent.

\paragraph{Frequent Appearance of Heart and Knife Emojis in SI Cases}  
The combination of \includegraphics[scale=0.05]{images/broken_heart.png} (broken heart) and \includegraphics[scale=0.05]{images/knife.png} (knife) appears in several high-frequency cases, such as \includegraphics[scale=0.05]{images/disappointed.png}\includegraphics[scale=0.05]{images/broken_heart.png}\includegraphics[scale=0.05]{images/knife.png}, \includegraphics[scale=0.12]{images/low_sad.jpg}\includegraphics[scale=0.05]{images/broken_heart.png}\includegraphics[scale=0.05]{images/knife.png}, \includegraphics[scale=0.05]{images/disappointed.png}\includegraphics[scale=0.05]{images/knife.png}\includegraphics[scale=0.05]{images/broken_heart.png}. This aligns with expressions of emotional pain, heartbreak, and themes of self-harm or suicidal thoughts. The juxtaposition of these symbols underlines the severity of distress felt by individuals in these situations, with the heart symbolizing emotional pain and the knife indicating self-harm tendencies.

\paragraph{Peaceful \& Hopeful Emojis Are Also Present}  
Despite the overwhelming presence of distressing symbols, some Serious Intent cases feature more hopeful or peaceful emojis. For example, \includegraphics[scale=0.15]{images/dove.png} (dove), \includegraphics[scale=0.05]{images/purple_heart.png} (purple heart), \includegraphics[scale=0.05]{images/sparkles.png} (sparkles), and \includegraphics[scale=0.05]{images/rain_cloud.png} (rainbow) appear in a subset of cases. This suggests that not all messages of distress are devoid of positive elements. Some users may incorporate symbols of support, comfort, and hope, as seen in combinations like \includegraphics[scale=0.05]{images/hugging_face.png}\includegraphics[scale=0.05]{images/purple_heart.png}\includegraphics[scale=0.15]{images/dove.png}, which possibly represent messages of encouragement or solidarity.

\paragraph{Lack of Casual Mentions in Most Cases}  
An interesting observation is that the top-ranked emoji combinations in Serious Intent cases consistently show zero Casual Mentions Count (CM). This indicates that users typically reserve these complex emoji combinations for serious situations, avoiding casual or lighthearted use. The seriousness of the emotional content in these cases is reinforced by the absence of trivial references.

\paragraph{Very Few Humor or Sarcasm-Oriented Emoji Combinations}  
While humor or sarcasm may be present in some datasets through emojis like \includegraphics[scale=0.013]{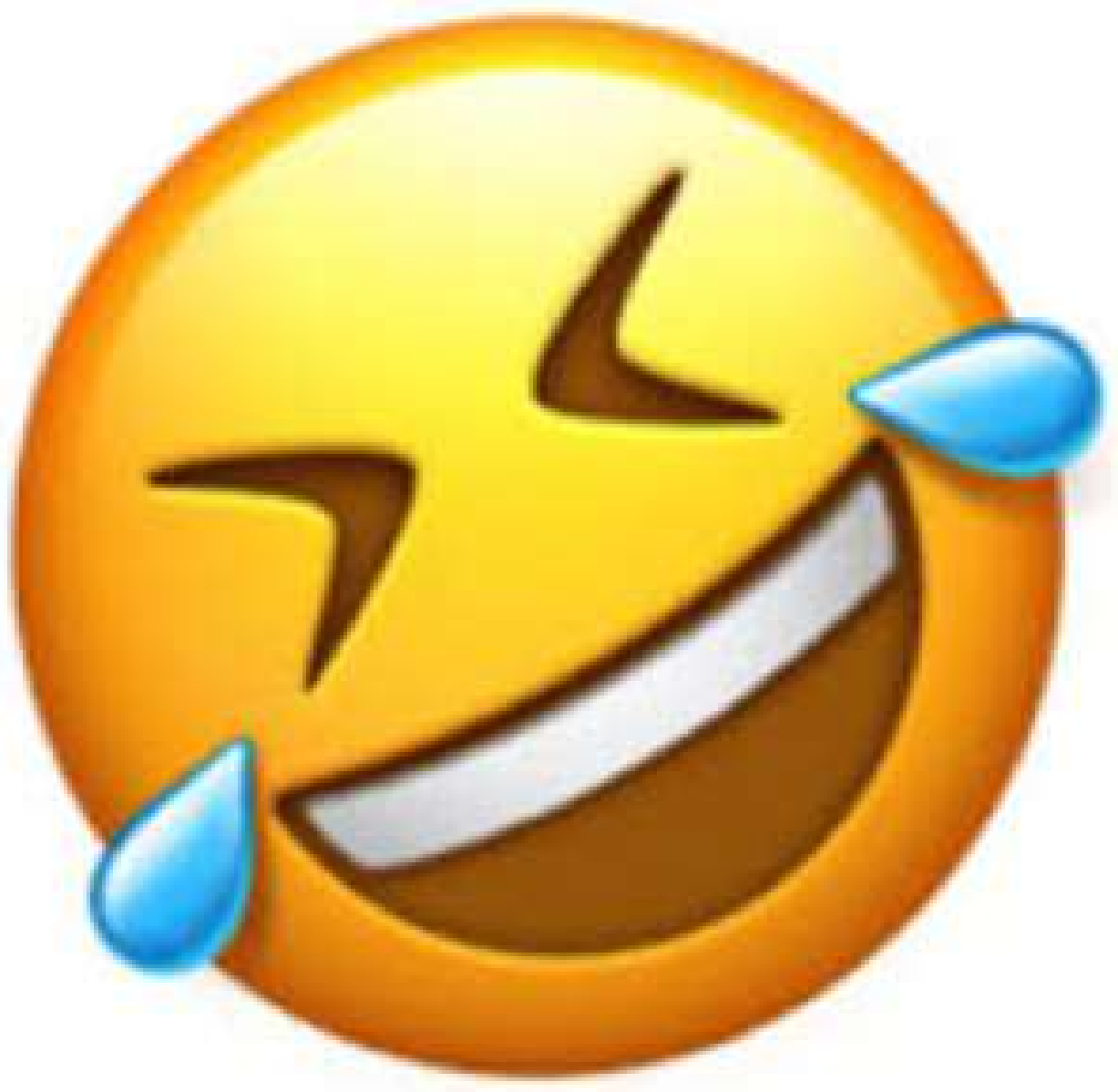} (laughing), \includegraphics[scale=0.05]{images/upside_down_face.png} (upside-down face), and \includegraphics[scale=0.05]{images/sweat_smile.png} (awkward smile), this dataset does not show significant occurrences of such patterns. This suggests that, in the context of Serious Intent, individuals are more likely to express their distress directly rather than masking it with dark humor or sarcasm.

\paragraph{Appearance of "Reinforcing" Emojis in Self-Harm Contexts}  
Some self-harm-related emoji combinations, such as \includegraphics[scale=0.05]{images/new_moon.png}\includegraphics[scale=0.05]{images/coffin.png}\includegraphics[scale=0.05]{images/disappointed.png}, \includegraphics[scale=0.05]{images/broken_heart.png}\includegraphics[scale=0.05]{images/knife.png}\includegraphics[scale=0.05]{images/rain_cloud.png}, and \includegraphics[scale=0.05]{images/pensive.png}\includegraphics[scale=0.05]{images/knife.png}\includegraphics[scale=0.05]{images/broken_heart.png}, include repeated symbols that reinforce the distressing emotions being conveyed. The use of multiple reinforcing emojis in sequence may indicate a stronger emotional expression, emphasizing the severity of the distress or self-harm ideation that the individual is experiencing.






\subsection{Experimental Setup: Additional Details}
We trained the model for 4 epochs using the AdamW optimizer (weight decay: 0.01) with a learning rate of 4e-5 and a linear decay scheduler, including a 10\% warm-up phase (500 steps). A batch size of 16 and sequence length of 256 tokens were used. Training was conducted over 4 epochs, with validation loss closely monitored to prevent overfitting. We used the AdamW optimizer with a weight decay of 0.01 to enhance generalization. A 0.2 dropout rate and gradient clipping at 1.0 were applied for stability and to avoid overfitting. Model evaluation was performed after each epoch, with early stopping activated if no improvement was observed after 3 epochs. 

\subsection{Information on the Evaluation Metrics}

\begin{itemize}
    \item \textit{F1 for Span Extraction:} We follow the calculation of F1 as done in \cite{poria2021recognizing}, which was inspired by the work of \citet{rajpurkar2016squad}. This metric measures the average overlap between the prediction and ground truth answer. We treat the prediction and ground truth as bags of tokens and compute their F1 score. We then take the maximum F1 over all of the ground truth answers for a given question and average this over all questions.
    \item \textbf{Relevance}
        \begin{itemize}[nolistsep]
            \item \textbf{Intuition}: This metric checks if the rationale includes all the key phrases related to casual mentions and serious intents.
            \item \textbf{Functionality}:
                \begin{itemize}[nolistsep]
                    \item Combines the casual mentions and serious intents into a single text.
                    \item Converts both the combined text and the rationale to lowercase.
                    \item Checks if all spans are present in the rationale.
                \end{itemize}
        \end{itemize}
    \item \textbf{Coherence}
        \begin{itemize}[nolistsep]
            \item \textbf{Intuition}: This metric evaluates the logical consistency and smoothness of the rationale in relation to the spans.
            \item \textbf{Functionality}:
                \begin{itemize}[nolistsep]
                    \item Combines the casual mentions and serious intents into a single text.
                    \item Uses TF-IDF Vectorizer to transform both the combined text and the rationale into vectors.
                    \item Computes cosine similarity between these vectors.
                \end{itemize}
        \end{itemize}
    \item \textbf{Readability}
        \begin{itemize}[nolistsep]
            \item \textbf{Intuition}: Measures how easy it is to read and understand the rationale.
            \item \textbf{Functionality}:
                \begin{itemize}[nolistsep]
                    \item Uses the Flesch-Kincaid grade level formula to compute the readability score of the rationale.
                \end{itemize}
        \end{itemize}
    \item \textbf{Semantic Similarity}
        \begin{itemize}[nolistsep]
            \item \textbf{Intuition}: Evaluates the conceptual similarity between rationale and spans.
            \item \textbf{Functionality}:
                \begin{itemize}[nolistsep]
                    \item Combines the casual mentions and serious intents into a single text.
                    \item Uses a pre-trained sentence embedding model (all-MiniLM-L6-v2)\footnote{\url{https://huggingface.co/sentence-transformers/all-MiniLM-L6-v2}} to generate embeddings for both the combined text and the rationale.
                    \item Computes cosine similarity between these embeddings.
                \end{itemize}
        \end{itemize}
\end{itemize}

\begin{table*}[h!]
\centering
\renewcommand{\arraystretch}{1}
\caption{Fine-tuning and Rationale Generation Prompts for our Proposed Setup.}\label{tab:Aprompt}
\begin{adjustbox}{max width=\textwidth}
\begin{tabular}{p{0.63\textwidth}|p{0.63\textwidth}}
\hline
\textbf{Fine-tuning Prompt} & \textbf{Rationale Generation Prompt} \\ \hline
\vspace{-3mm}
\begin{lstlisting}[escapeinside={(*}{*)}]
"instruction": "Analyze the following social media post 
to determine if it is related to self-harm. Identify spans 
related to casual mention and serious intent regarding self-
harm.",
"input": {
    "post text": "I'm really struggling right now. 
    Everything feels pointless. I've thought about cutting again, 
    just to feel something. (* \includegraphics[scale=0.05]{images/disappointed.png} *)",
    "emojis": [
        {
            "emoji": " (* \includegraphics[scale=0.05]{images/disappointed.png} *)",
            "usual_meaning": "Disappointed Face",
            "contextual_meaning": "Sadness",
            "casual mention chance": "Medium",
            "serious intent chance": "High"
        }
    ]
},
"output": {
    "classification": "self-harm",
    "casual_mention_spans": [],
    "serious_intent_spans": ["thought about cutting again"]
}
\end{lstlisting} \vspace{-7mm}&
\vspace{-3mm}
\begin{lstlisting}[escapeinside={(*}{*)}]
"instruction": "Using the provided classification and 
extracted spans, generate a rationale explaining why the post 
is classified as self-harm or non-self-harm. Consider the 
spans and the emoji meanings provided.",
"input": {
    "post text": "I'm really struggling right now. 
    Everything feels pointless. I've thought about cutting again, 
    just to feel something.  (* \includegraphics[scale=0.05]{images/disappointed.png} *)",
    "classification": "self-harm",
    "casual_mention_spans": [],
    "serious_intent_spans": ["thought about cutting again"],
    "emojis": [
        {
            "emoji": " (* \includegraphics[scale=0.05]{images/disappointed.png} *)",
            "usual_meaning": "Disappointed Face",
            "contextual_meaning": "Sadness",
            "casual mention chance": "Medium",
            "serious intent chance": "High"
        }
    ]
},
"output": ""
\end{lstlisting} \vspace{-7mm} \\ \hline
\end{tabular}
\end{adjustbox}
\end{table*}

\begin{table*}[!ht]
\centering
\caption{Zero-Shot, Few-Shot, and Fine-Tuning Prompts for Self-Harm Classification and Rationale Generation}
\label{tab:prompts}
\renewcommand{\arraystretch}{1.5}
\begin{tabular}{l|l}
\hline
\textbf{Type} & \textbf{Prompt} \\
\hline
\small \textbf{Zero-Shot} &
\adjustbox{max width=\textwidth}{
\begin{minipage}[t]{0.8\textwidth}
\small \textbf{\#\# Instruction}

\small You are tasked with evaluating social media posts to classify them as self-harm or non self-harm. Review the post below and provide your classification. Additionally, write a detailed rationale that explains your decision, taking into account the content and context of the post. \\

\small \textbf{\#\# Input}

\small \{post text\}\\

\small \textbf{\#\# Output}

\small \textbf{Classification}: [self-harm / non self-harm] \\
\small \textbf{Rationale}: {Describes the factors and considerations that guided the classification decision.}\\
\end{minipage}} \\
\hline

\small \textbf{Few-Shot} &
\adjustbox{max width=\textwidth}{
\begin{minipage}[t]{0.8\textwidth}
\small \textbf{\#\# Instruction}

You are tasked with evaluating social media posts to classify them as self-harm or non self-harm. Review the following posts and classifications to understand the task. Then classify the new post and write a detailed rationale that explains your decision, taking into account the content and context of the post. \\

\textbf{\#\# Input}

\textbf{Example 1}: \\
\textbf{Post}: ``I feel so worthless. Sometimes I think about hurting myself to make the pain stop." \\
\textbf{Classification}: self-harm \\
\textbf{Rationale}: The mention of ``hurting myself" suggests a serious consideration of self-harm. \\
\textbf{Example 2}: \\
\textbf{Post}: ``Just had a really tough day at work, but I’m trying to stay positive!" \\
\textbf{Classification}: non self-harm \\
\textbf{Rationale}: The post expresses frustration but lacks any direct mention of self-harm. \\
\textbf{New Post}: \\
\{new post text\}\\

\textbf{\#\# Output}

\textbf{Classification}: [self-harm / non self-harm] \\
\textbf{Rationale}: {Describes the factors and considerations that guided the classification decision.}\\

\end{minipage}} \\
\hline

\small \textbf{Fine-Tuning} &
\adjustbox{max width=\textwidth}{
\begin{minipage}[t]{0.8\textwidth}
\small The single-task fine-tuning prompt for self-harm classification and the subsequent prompt for rationale generation are similar to the one depicted in Table~\ref{tab:Aprompt}, with the exception that they do not include CM and SI spans, as well as emoji information sourced from the CESM-100 dataset.\\
\end{minipage}} \\
\hline

\end{tabular}
\end{table*}

By employing these metrics, we can ensure that the generated rationales for self-harm detection are accurate, consistent, readable, and semantically aligned with the extracted spans.

\subsection{Model Details}\label{modeldetails}
\begin{itemize}
    \item \textbf{Mental-Alpaca-7B \cite{xu2024mental}:} An LLM fine-tuned for mental health prediction using Alpaca, trained on diverse datasets including Dreaddit, DepSeverity, SDCNL, and CCRS-Suicide for enhanced accuracy in mental health analysis.
    \item \textbf{Llama-3.1-8B-Instruct \cite{llama3modelcard}:} Developed by Meta Inc., Llama 3 is a state-of-the-art LLM outperforming many open-source models on common benchmarks. We use the pre-trained 8B version for our experiments.
    \item \textbf{MentaLLaMA-chat-7B \cite{yang2024mentallama}:} Specialized Llama2-chat-7B model fine-tuned on the IMHI dataset for interpretable mental health analysis, offering reliable explanations and state-of-the-art condition prediction.
\end{itemize}
All models are sourced from Hugging Face\footnote{\url{https://huggingface.co/models}}.

\subsection{Prompts}\label{app:prompts}
Table~\ref{tab:Aprompt} presents the fine-tuning and rationale generation prompts for our proposed setup. We present the prompts for the various baseline setups in Table~\ref{tab:prompts}. We strategically selected few-shot examples to ensure a balanced and diverse representation of self-harm classification cases. This included two Casual Mention (CM) examples, two Serious Intent (SI) examples, and one borderline case with ambiguous intent. To enhance robustness, we conducted three runs with different random sets and had human annotators verify the representativeness of the selected samples. We present the prompts used for generating the synthetic data in Table~\ref{tab:prompts for synthetic data}.

\subsection{Error analysis}\label{App:errana}
\begin{itemize}
    \item \textit{Ambiguity in Recovery Language:} Posts containing both self-harm references and hopeful or recovery-oriented language may lead to mixed classifications. Models like Mental-Alpaca and MentalLlama sometimes misclassify posts as non self-harm due to a focus on positive language and recovery hashtags, overlooking the serious self-harm content.

    \item \textit{Current vs. Past Intent:} Distinguishing between current and past self-harm intent is crucial. While all models correctly handle past ideation, the challenge arises in posts where past struggles are intertwined with current emotional states. Clearer delineation in models' handling of past versus present intent could improve accuracy.
\end{itemize}

\begin{table*}[!ht]
\centering
\caption{Prompts for generating synthetic self-harm and non-self-harm samples in our dataset}
\label{tab:prompts for synthetic data}
\renewcommand{\arraystretch}{0.8}
\begin{tabular}{p{2cm}|p{12.8cm}}
\hline
\textbf{\small Label of synthetic post} & \textbf{Prompt} \\
\hline
\textbf{\small self-harm} &
\textbf{\small \#\# Instruction}

\small Create a Reddit post where the author reflects on their experiences with self-harm. The post should express a real person's emotions and thoughts, incorporating past incidents, current struggles, or stories related to their journey. Encourage the author to discuss their feelings, the impact on their life, or how they’ve tried to cope with their condition. The language should be varied and natural, avoiding repetition. \\
& \\
& \textbf{\small \#\# Input}

\textbf{\small Example 1}: \\
& \textbf{\small Post}: \small ``I keep finding myself tracing the scars on my skin, a reminder of the war inside my head. The urge to self-harm is like a shadow that never leaves, lurking in the corners of my mind. It's a battle between wanting to feel something and wanting the pain to stop. Sometimes the blade feels like the only friend who understands. But deep down, I know it's not the answer. Trying to hold on to hope like a fragile thread in a storm. One day at a time, one breath at a time. We're warriors fighting invisible battles, and our scars tell stories of survival." \\
& \small \textbf{serious intent spans}: ["tracing the scars on my skin",``urge to self-harm",``the blade feels like the only friend"] \\
& \small \textbf{Rationale}: The reference of ``urge to self-harm" and ``the blade feels like the only friend" is unquestionably an indication of the author's genuine resolve to cause physical harm to themselves.\\
\\
& \small \textbf{Example 2}: \\
& \small \textbf{Post}: ``It’s been years since I last hurt myself, but yesterday, I had the strongest urge to do it again. I found an old blade in my drawer and just sat there holding it, trying to convince myself not to use it. I didn’t, but it was so hard." \\
& \small \textbf{serious intent spans}: ["since I last hurt myself",``strongest urge to do it again"] \\
& \small \textbf{Rationale}: The use of phrases like ``strongest urge to do it again" by the user exhibit a definite intention of causing self-injury, combined with the context of the post.  \\
& \\
& \small \textbf{\#\# Output}

 \\
\hline

\small \textbf{Non-self harm}

& \small \textbf{\#\# Instruction}

\small Create a Reddit post containing phrases which indicate self harm in a joking or casual way. Phrases will be containing the self-harm intentions in a sarcastic way where the author is not actually going to harm himself or herself. It can show the irritation, stress, anger or disgrace just to express the situation. The author should be in a light mood to express his/her thoughts. \\
& \\
& \small \textbf{\#\# Input}\\

& \small \textbf{Example 1}: \\
& \small \textbf{Post}: ``Great, my boss just gave me another pointless task. Guess I'll just bang my head against the wall." \\
& \small \textbf{Casual intent spans}: ["bang my head against the wall"] \\
& \small \textbf{Rationale}: The mention of ``bang my head against the wall" is a violent expression suggesting intention to hurt themselves, but,with the context of the situation, it is clearly just a metaphorical expression for frustration and does not show an actual desire for self harm. \\ \\
& \small \textbf{Example 2}: \\
& \small \textbf{Post}: ``Lost all my progress because of a glitch. Might as well just throw myself off a cliff." \\
& \small \textbf{Casual intent spans}: [" throw myself off a cliff."] \\
& \small \textbf{Rationale}: The mention of ``throw myself off a cliff" shows exasperation of author about loosing his work, but it is clear looking at the emojis that it is definitely not said in a serious intent of harming themselves. \\

& \\
& \small \textbf{\#\# Output}
 \\
\hline

\end{tabular}
\end{table*}

\end{document}